\pdfoutput=1
\documentclass[a4paper, 10pt, conference]{ieeeconf}      
\usepackage{FG2026}

\usepackage{colortbl}
\definecolor{iccvblue}{rgb}{0.21,0.49,0.74}
\usepackage[pagebackref,breaklinks,colorlinks,allcolors=iccvblue]{hyperref}
\usepackage{booktabs}
\usepackage{subcaption}
\usepackage{multirow}

\FGfinalcopy 

\IEEEoverridecommandlockouts                              
\usepackage{times} 
\usepackage{amsmath} 
\usepackage{amssymb}  

\title{\LARGE \bf
Self-Learning Expression Deformations for Data-Efficient\\Gaussian Avatars
}
\author{Jiahao Yang \;\;\;\; Xiaohang Yang \;\;\;\; Qing Wang \;\;\;\; Yilan Dong \;\;\;\; Gregory Slabaugh \;\;\;\; Shanxin Yuan\\
Queen Mary University of London\\
}
\begin{document}
\pagestyle{plain}
\thispagestyle{plain}

\twocolumn[{
\renewcommand\twocolumn[1][]{#1}
\maketitle
\begin{center}
    \centering
    \vspace{-0.5cm}
    \captionsetup{type=figure}
    \includegraphics[width=0.9\linewidth]{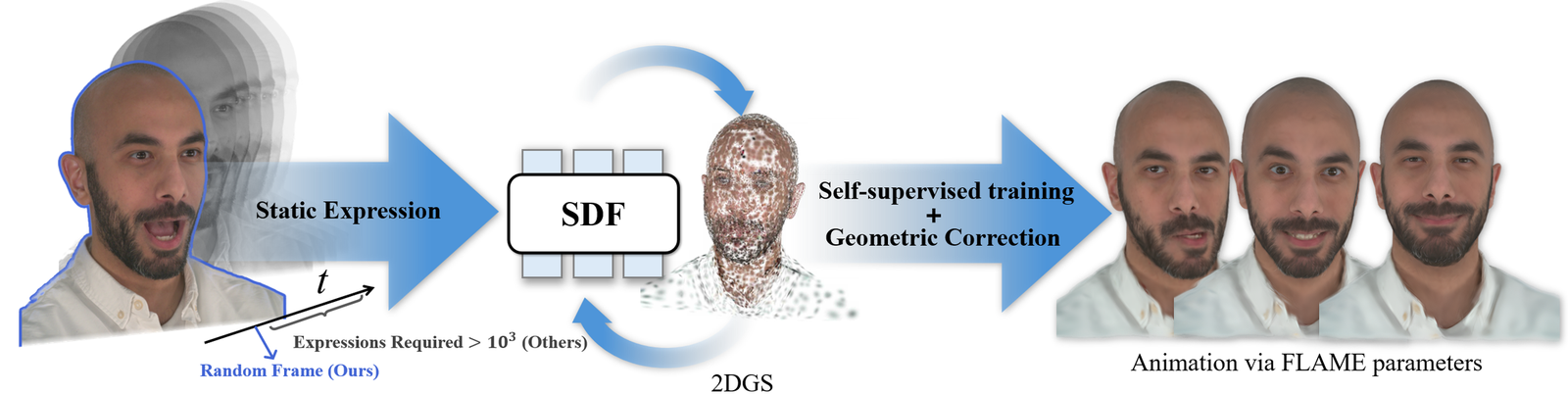}
    \captionof{figure}{We introduce SAGE, a method for creating animatable Gaussian avatars from minimal training data. Our pipeline consists of two training phases: (1) joint optimization of an SDF and 2DGS for accurate surface reconstruction, and (2) a self-supervised strategy to ensure appearance and geometric robustness under deformation. Additionally, we employ refinement of Gaussian shapes using surface properties during animation. Finally, the avatar is animated using FLAME parameters.}
    \label{fig:idea}
\end{center}
}]

\begin{abstract}
Modeling dynamic facial expressions using 3D Gaussian representations remains challenging due to their unstructured nature. Conventional Gaussian avatar pipelines require extensive multiview and sequential expression data, limiting scalability and accessibility. In this work, we introduce \textbf{S}elf-\textbf{A}daptive \textbf{G}aussian \textbf{E}xpression (SAGE), a framework for self-learning expression-induced Gaussian deformations that enables high-fidelity, animatable avatars from minimal input data. Our method jointly optimizes 2D Gaussian surfels and a Signed Distance Field (SDF) to enforce compact, surface-aligned Gaussian distributions, while a self-supervised expression learning phase replaces long training sequences with geometric and appearance consistency constraints. This design allows flexible deployment across multiple reconstruction regimes: in the multiview setting, only a single frame (timestep) is required instead of thousands; in the monocular setting, only head rotations are needed without expression sequences; and in the one-shot setting, no pretraining or priors are necessary. Experiments demonstrate that our approach achieves reconstruction and animation quality comparable to state-of-the-art methods, while reducing data requirements by several orders of magnitude. Our results highlight the potential of self-supervised Gaussian deformation learning as a step toward accessible, data-efficient avatar creation.
\end{abstract}
\section{Introduction}
Creating photorealistic 3D avatars is a key challenge in computer vision, with applications in film, gaming, and animation. Facial modeling remains complex due to the need to capture appearance, geometry, and dynamics within a unified framework. Meanwhile, the demand for real-time rendering on lower-end devices drives the need for efficient, lightweight animation pipelines.

Recently, advancements in 3D Gaussian Splatting (3DGS) \cite{kerbl_3d_2023,charatan_pixelsplat_2024, chen_mvsplat_2025, hu_gauhuman_2023, yu_mip-splatting_2023} have outperformed implicit neural representations and inherited key techniques from prior advancements. 3D Morphable Model (3DMM)-based approaches remain preferable for creating Gaussian avatars \cite{xu_gaussian_2024, qian_gaussianavatars_2024, giebenhain_npga_2024, xiang_flashavatar_2024, ma_3d_2024}. However, 3DGS representations lack inherent structure, often introducing artifacts when deformed. Hence, these pipelines heavily rely on extensive sequential expression datasets \cite{kirschstein_nersemble_2023, zheng_ilsh_2023, pan_renderme-360_2023}, which are costly and unsuitable for wider use cases. Efforts in few-shot avatar creation \cite{deng_portrait4d_2024, chen_morphable_2024, zheng_headgap_2025, zielonka_synthetic_2025, yu_one2avatar_2024} still require extensive pre-training, leaving no efficient pipeline for avatar creation from minimal input.

To address this gap, we propose replacing long training sequences with geometric constraints to ensure consistency in Gaussian transformation, thereby enabling an efficient and accessible pipeline. We introduce SAGE, a framework for self-learning expression-induced Gaussian deformation. This allows us to achieve the following: reconstructing an avatar from a single frame\footnote{In this paper, we refer to a `single frame' as a set of multiple images captured from different viewpoints at a single timestep.} (timestep) of images of multiple viewpoints; reconstruction from monocular video, which only includes sufficient head rotation angles; and one-shot reconstruction using only one image. During optimization, we follow \cite{qian_gaussianavatars_2024}, rigging Gaussians to a FLAME \cite{li_learning_2017} mesh for direct control via FLAME parameters. 

The unstructured nature of Gaussian representations causes geometric and visual distortions in dynamic scenes. Inspired by 3DGS-based surface reconstruction methods \cite{huang_2d_2024,lyu_3dgsr_2024,chen_neusg_2023}, our approach jointly refines a Signed Distance Field (SDF), which serves as surface supervision to orderly align Gaussians. Additionally, we adopt 2D Gaussian Splatting (2DGS)'s \cite{huang_2d_2024} surfel representation instead and impose constraints to maintain a compact, tangential distribution. In addition, we devise separate networks to parametrize Gaussian appearances, reducing local Gaussian shape deviations. Inspired by geometry processing from \cite{yang_geometry_2021}, we utilize both the surface gradient and Hessian matrix to infer surface stretching and curvature to make adjustments to Gaussian shapes ensuring consistency during animation time.

Finally, we introduce a novel self-supervision strategy and train the avatar with expression parameter sequences, but without ground truth images. We use depth and normal constraints to ensure Gaussians remain compact and tangential to deformed surfaces. Additionally, appearance supervision is applied by comparing corresponding query points rendered by Gaussian splatting and a canonical appearance network. Experiments indicate our pipeline achieves performance comparable to state-of-the-art methods trained on thousands of frames, significantly reducing optimization time while maintaining robustness across diverse transformations.  

Overall, our contributions are as follows:  
\begin{itemize}
    \item We propose a strategy for generating surface-aligned, geometry-aware Gaussians that improve spatial consistency and alignment across deformations while maintaining visual accuracy.
    \item To enhance transformation robustness, we introduce a novel self-supervised training approach for fine-tuning Gaussian shape.  
    \item The established training strategy reduces computational costs and data requirements exponentially while achieving results comparable to state-of-the-art in each reconstruction modes.
      
\end{itemize}
\section{Related Work}
\label{sec:formatting}

\noindent\textbf{Animatable Head reconstruction} Early works on human head reconstruction relied on 3DMMs \cite{blanz_morphable_1999, cao_facewarehouse_2014, li_learning_2017, pavlakos_expressive_2019}, which parameterize shape and texture in low-dimensional spaces. Neural extensions \cite{giebenhain_learning_2023, yenamandra_i3dmm_2020, zheng_imface_2022} improve expression and geometry by learning residuals on top of mesh templates such as FLAME \cite{moon_authentic_2024, wang_faceverse_2022}, while others integrate neural textures \cite{grassal_neural_2022, kim_deep_2018, ma_pixel_2021, tewari_mofa_2017, feng_learning_2021}. More recent efforts shift to neural fields \cite{athar_rignerf_2022, gafni_dynamic_2020, gao_reconstructing_2022, hong_headnerf_2022, lombardi_mixture_2021, xu_avatarmav_2023, kirschstein_diffusionavatars_2024} and point-based methods such as Gaussian Splatting \cite{chen_monogaussianavatar_2023, qian_gaussianavatars_2024, xu_gaussian_2024, zheng_pointavatar_2023}. Expression-conditioned NeRFs leverage 3DMM coefficients, hash tables, or motion voxel grids \cite{gafni_dynamic_2020, gao_reconstructing_2022, xu_avatarmav_2023}, while implicit deformation networks \cite{park_nerfies_2021, park_hypernerf_2021, pumarola_d-nerf_2020} are extended with expression, audio, or FLAME parameters \cite{yan_dialoguenerf_2023, wang_morf_2022, athar_rignerf_2022, zhuang_mofanerf_2022}. Explicit approaches instead impose geometric consistency, e.g., aligning surface points to a canonical mesh \cite{zielonka_instant_2023} or applying FLAME-based Linear Blend Skinning \cite{zheng_i_2022, zheng_pointavatar_2023}. For 3DGS, controllability can be introduced via tetrahedral cages \cite{zielonka_drivable_2023}, SMPL with shell maps \cite{jena_splatarmor_2023}, or blendshape-driven LBS \cite{ma_3d_2024}. Robustness is further improved with per-Gaussian parameters \cite{dhamo_headgas_2023, zheng_gps-gaussian_2024, wang_gaussianhead_2024}, mesh-bound Gaussians \cite{qian_gaussianavatars_2024}, or expression-conditioned MLPs \cite{xu_gaussian_2024, giebenhain_npga_2024, xiang_flashavatar_2024, teotia_gaussianheads_2024}. But these pipelines heavily rely on extensive sequential datasets to achieve generalizability for novel expression reconstruction.

\noindent\textbf{Few-Shot Head Avatar} Few-shot creation methods span several directions. \cite{kirschstein_gghead_2024, lyu_facelift_2025} can generate high fidelity Gaussian avatar, but does not support animation. Latent deformation encodings with 2D generative models enable facial reenactment from limited inputs~\cite{ren_pirenderer_2021, wiles_x2face_2018, siarohin_first_2020}. StyleGAN2~\cite{karras_analyzing_2020} and its latent space are frequently adapted, often combined with monocular 3DMM tracking to improve 3D geometry, pose consistency, and controllable animation. Another line of work learns 3D head priors from large-scale data for personalization: Morphable Diffusion~\cite{chen_morphable_2024} applies diffusion to 3DMM meshes for single-image avatar creation. VAE-based frameworks~\cite{moon_authentic_2024, buhler_preface_2023, yu_one2avatar_2024, yang_vrmm_2024} encode appearance and adapt head priors to new identities; \cite{zheng_headgap_2025, saunders_gasp_2024, xu_gphm_2024, zielonka_synthetic_2025, he_lam_2025} built a generalizable Gaussian prior learned from large datasets to enable few-shot avatar creation. While such generative approaches achieve strong results, they often require extensive pre-training. By contrast, our method is purely geometric and only requires minimal data for training.

\noindent\textbf{Surface Geometry of 3DGS} Recent approaches have attempted to reconstruct accurate surfaces via 3DGS. SuGaR \cite{guedon_sugar_2023} approximates Gaussians with near planar primitives and binary opacity. NeuSG \cite{chen_neusg_2023} integrates 3DGS with NeuS \cite{wang_neus_2023}, encouraging Gaussians to align more compactly with an underlying SDF. 3DGSR \cite{lyu_3dgsr_2024} conditions Gaussian opacities to nearest‐surface distances via radial basis kernels, adding depth/normal cues from \cite{gao_relightable_2023}. 2DGS \cite{huang_2d_2024} reduces 3D ellipsoid Gaussians to planar discs for better surface alignment, \cite{lee_surfhead_2025, chen2025mixedgaussianavatarrealisticallygeometricallyaccurate, schoneveld2025sheapselfsupervisedheadgeometry} have later adapted 2DGS to avatar creation.  Our work follows the above geometric approach, optimizing Gaussian primitives to align tangentially to the underlying surface. The resulting Gaussian achieves view and transformation consistent, while only trained using one frame of multi-view data, in contrast with current works which takes thousands of frames to train.

\begin{figure*}[t]
    \centering
    \includegraphics[width=\linewidth]{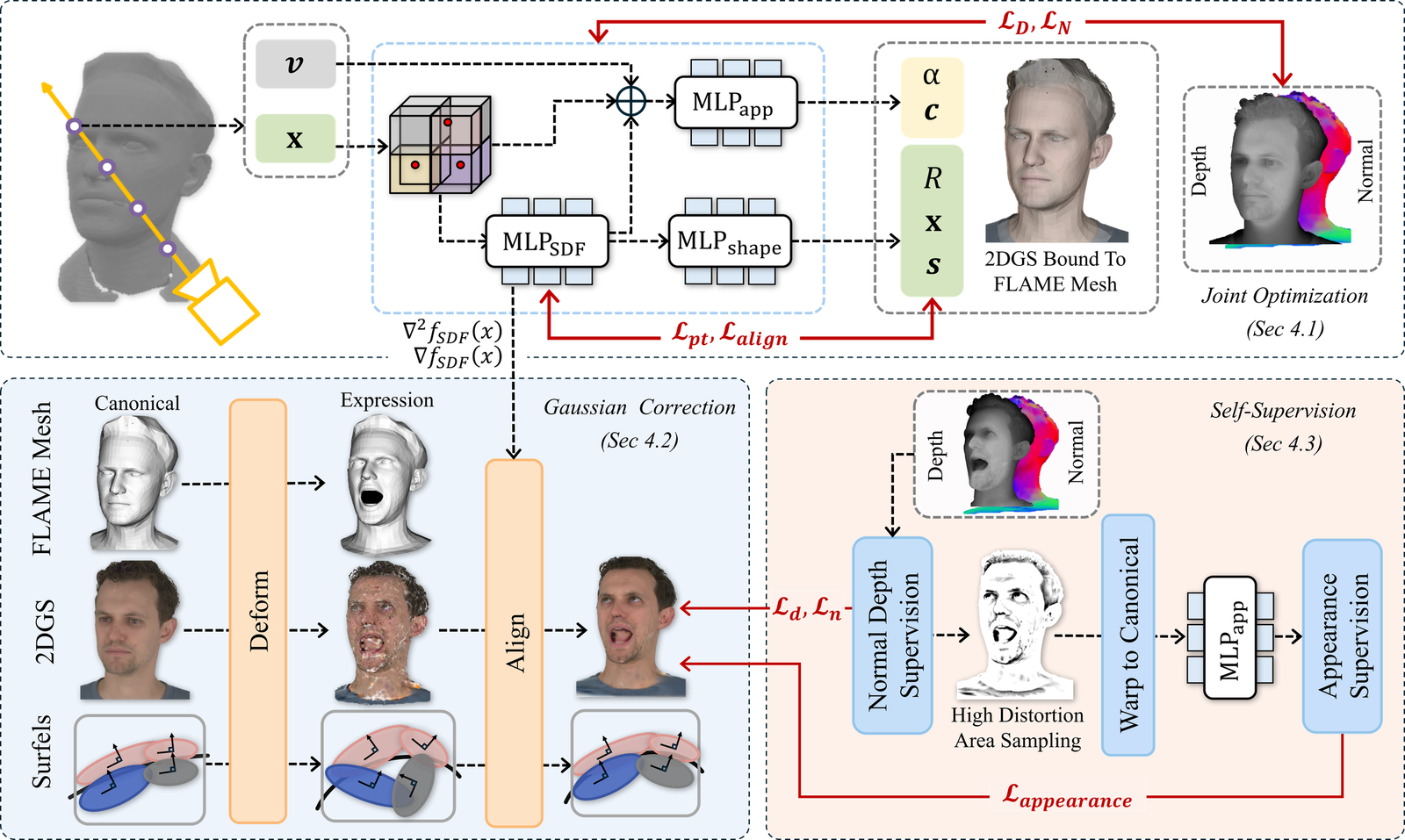}
    \caption{Joint Optimization (top): SDF branch predicts Gaussian attributes and 2DGS will rasterize both depth and normal map, for supervising the SDF. SAGE's adaptation in a dynamic scene (bottom): Gaussian shape correction is illustrated on the left, when our avatar undergo expression changes, principal curvature and stretch is approximated to adjust $\{\bm{R},\bm{s}\}$. On the right is the second training phase, during which we optimize $\mathcal{L}_n$ and $\mathcal{L}_d$, regions of low consistency and distortion are sampled to compare their appearance in canonical and deformed pose.}
    \label{fig:enter-label}
\end{figure*}
\section{Preliminaries}
\textbf{3DGS} \cite{abdal_gaussian_2023} is a point-based representation, consists of Gaussians parametrized by position $\textbf{x}$, rotation $\bm{R}$, scale $\bm{s}$, opacity $\alpha$, color $\bm{c}$ and spherical harmonics $\bm{sh}$.
\begin{equation}
    \mathcal{G} = \{\textbf{x}, \bm{R}, \bm{s}, \alpha, \bm{c}, \bm{sh}\}, \quad I = \mathcal{\text{Rasterize}}(\mathcal{G}, \pi_{\mathit{K}, \mathit{E}}).
     \label{eq:rasterization}
\end{equation}
Here we use $\mathcal{G}$ to denote attributes parametrizing Gaussians which are used to render an image $I$ using tile-based rasterization given intrinsic and extrinsic camera parameters $\pi_{\mathit{K}}$ and $\pi_{\mathit{E}}$, respectively.

\noindent\textbf{2DGS} \cite{huang_2d_2024} on the other hand, replaces elliptical Gaussians with 2D discs to improve alignment with thin surfaces. Consequently,  $\bm{R}$ and $\bm{s}$ from Eq. \eqref{eq:rasterization} are reduced to two tangent vectors: \(\mathbf{t}_1, \mathbf{t}_2 \)  and two scaling factors \( s = (s_1, s_2) \) controlling Gaussian shape, with the normal of the surfel given by \( \mathbf{t}_3 = \mathbf{t}_1 \times \mathbf{t}_2 \). Typically, approximating 3D Gaussians with their centers and performing alpha blending only yields a pseudo depth map. 2DGS allows for precise depth calculation at an intersection point. To enforce geometric consistency, the \textit{depth distortion loss} reduces depth variations among overlapping splats:
\begin{equation}
\mathcal{L}_d = \sum_{i,j} \omega_i \omega_j |z_i - z_j|_1,
\end{equation}
where \( z_i \) is the depth of the \( i \)-th splat intersection and \( \omega_i \) its blending weight. The \textit{normal consistency loss} aligns splat normals with depth-estimated surface normals:

\begin{equation}
\mathcal{L}_n = \sum_{i} \alpha_i (1 - |\mathbf{n}_i^\top \mathbf{N}|_1),
\end{equation}
where \( \mathbf{n}_i \) is the splat normal, \( \mathbf{N} \) the estimated surface normal, and \( \alpha_i \) the blending weight. These techniques enable efficient CUDA-based rendering and stable optimization across diverse transformations.

\noindent\textbf{GaussianAvatars} \cite{qian_gaussianavatars_2024} bind Gaussians to the tracked meshes and enable control via FLAME parameters. A local coordinate system is defined for each triangle in global space, with the mean vertex position \( \mathbf{o}_\mathcal{F} \) as the origin. The orientation is defined by \( \mathbf{R}_\mathcal{F} \), constructed from an edge direction, the triangle normal, and their cross product. The mean edge length \( k_\mathcal{F} \) normalizes local coordinates. Given a Gaussian’s local position \( \mathbf{x}^l \), local rotation $\bm{R}^l$, and scaling $\bm{s}^l$, their global space correspondences are: 
\begin{equation}
\mathbf{R}^G = \mathbf{R}_\mathcal{F} \mathbf{R}^l, \quad  \mathbf{x}^G = k_\mathcal{F} \mathbf{R}_\mathcal{F} \mathbf{x}^l + \mathbf{o}_\mathcal{F}, \quad \bm{s}^G = k_\mathcal{F} \bm{s}^l.
\end{equation}

At rendering time, the mesh is deformed using linear blend skinning (LBS) and FLAME parameters $\bm{\beta}, \bm{\theta}, \bm{\psi}$ and $\{\mathbf{o}_\mathcal{F}, \mathbf{R}_\mathcal{F},k_\mathcal{F} \}$ is updated to obtain global Gaussian positions at a new time instance:
\begin{equation}
    \{o_\mathcal{F}, R_\mathcal{F},k_\mathcal{F}\} = \mathcal{M}(\mathcal{F}, \bm{\beta}, \bm{\theta}, \bm{\psi}),
    \label{eq:flame_lbs}
\end{equation}
where M denotes the computation for performing LBS and obtaining $\{\mathbf{o}_F, \mathbf{R}_F,k_F\}$ from mesh $\mathcal{F}$ .
\section{Methods}
\textbf{Overview:}
Our pipeline integrates FLAME rigging with 2DGS to establish a foundation for reconstructing a geometry-accurate and controllable avatar. To extend consistency from static to dynamic scenarios, we introduce a two-phase training strategy and a Gaussian correction mechanism. The first training phase focuses on reconstructing an accurate neutral head representation. We parametrize the Gaussian shape and appearance attributes on two multi-layer perceptrons (MLP) separately. During animation, we utilize geometric properties from the SDF to perform Gaussian shape adjustments. Finally, we propose a novel self-supervised training strategy in the second phase to ensure attributes' robustness in dynamic scenes.

\subsection{Neural Surface Reconstruction} 
An SDF $f_{SDF} $ and Gaussian surfels are jointly optimized during the first training phase. We constrain the distance between Gaussians from the surface by minimizing the \( L_1 \)-norm of the SDF value at each Gaussian center, encouraging Gaussian primitives to remain close to the surface while providing supervision to the neural surface, ensuring convergence.

\begin{equation}
\mathcal{L}_{\text{pt}} = | f_{\text{SDF}}(\mathbf{x}^G) |_1.
\end{equation}

To further improve surface alignment, we enforce Gaussian surfels to lie tangential to the surface. Since the SDF gradient at the surface corresponds to the surface normal, we define an alignment loss that minimizes the angular deviation between the Gaussian normal $\mathbf{t}_3$ and the SDF-derived normal.

\begin{equation}\label{eq:7}
\mathcal{L}_{\text{align}} = \left| 1 - | {\mathbf{t}_3 ^G}^ \top \cdot \nabla f_{\text{SDF}}(\mathbf{x}^G) | \right|_1.
\end{equation}

Following MonoSDF \cite{yu_monosdf_2022}'s training strategy, we make use of the depth and normal maps rendered from 2DGS to supervise the SDF's volumetric renderings. Additionally, we adapt a smoothing constraint to penalize noisy regions resulting from  Gaussian over-cluttering, which is problematic in 2DGS's optimization.

\subsubsection*{Parametrized Gaussian Attributes }
As the Gaussian representation is inherently discrete, gradients do not spread to nearby primitives during backpropagation, and therefore individual Gaussians are uncorrelated. This leads to randomized attributes $\mathcal{G}$ and results in transformation inconsistency. To ensure our avatars reconstructed from static scenes remain consistent during deformation, we parameterize $\mathcal{G}$ using separate MLPs. Specifically, we condition $f_{app}$ on $f_{SDF}$ to predict Gaussian appearance attributes $\{\bm{c}, \alpha \}$ given query point $\mathbf{x}^G$ in global space:

\begin{equation}
    \hat{\mathbf{c}}, \hat\alpha = f_{app}(\mathbf{x}^G, \mathbf{v}, \hat{\mathbf{n}}, \hat{\mathbf{z}}),
\end{equation}

where $\mathbf{v}$ is the view direction, $\hat{\mathbf{n}}$ is the SDF gradient and $\hat{\mathbf{z}}$ is a feature vector provided by $f_{SDF}$ and querying $\{\Phi^{l}\}_l$. Together with depth and normal, the appearance network is regulated with volumetric rendering, for simplicity we use $\hat{C}(\mathbf{r})$, $\hat{D}(\mathbf{r})$, $\hat{N}(\mathbf{r})$ to represent the SDF volumetric rendering given ray $\mathbf{r}$:
\small\begin{align}
\mathcal{L}_{\text{rgb}} &= \sum_{\mathbf{r} \in R} | \hat{\mathbf{C}}(\mathbf{r}) - \mathbf{C}(\mathbf{r}) |^2, \\
\mathcal{L}_{\text{D}} &= \sum_{\mathbf{r} \in R} | \hat{D}(\mathbf{r}) - \hat D_{2D}(\mathbf{r}) |^2, \\
\mathcal{L}_{\mathbf{N}} &= \sum_{r \in \mathcal{R}} |\hat{\mathbf{N}}(\mathbf{r}) - \hat{\mathbf{N}}_{2D}(\mathbf{r})|_1 + |\mathbf{1} - \hat{\mathbf{N}}(\mathbf{r})^\top \hat{\mathbf{N}}_{2D}(\mathbf{r})|_1,
\end{align}
where $\hat{D}_{2D}(r)$, $\hat{N}_{2D}(r)$ represent render results from 2DGS. Therefore, the total loss for all MLPs is defined by:
\begin{equation}
\mathcal{L}_{sdf} = \mathcal{L}_{\text{rgb}} + \lambda_{\text{pt}} \mathcal{L}_{\text{pt}} + \lambda_{\text{align}}\mathcal{L}_{\text{align}}+ \lambda_{\text{D}} \mathcal{L}_{\text{D}} + \lambda_{\text{N}} \mathcal{L}_{\text{N}},
\end{equation}
and \( \lambda_{\text{pt}}, \lambda_{\text{align}}, \lambda_{\text{D}}, \lambda_{\text{N}} \) are weighting terms controlling the impact of each loss component.

Furthermore, we make use of the multi-resolution features stored in the hash grid, by conditioning on the queried feature, a network $f_{shape}$ is used to predict $\{\bm{R}^G, \bm{s}^G\}$:
\begin{equation}
    \{\bm{R}^G, \bm{s}^G\} = f_{shape} \left(\mathbf{x}^G, \left\{ \textit{interp}(\mathbf{x}^G, \Phi^{l}) \right\}_{l} \right),
\end{equation}
where $\textit{interp}( \cdot )$ represents the interpolation procedure for querying feature grid $\Phi$ at resolution level $l$.

\subsection{Gaussian Surfel Correction}
\label{sec:correction}
We propose a novel approach to maintain surface integrity during deformation by adjusting Gaussian attributes based on estimated surface stretching and curvature. For each surfel (illustrated in Fig. \ref{fig:enter-label}), we determine its \( k \)-nearest neighbors via ball query in a small radius based on the scale of the surfel's bound mesh triangle. We approximate geometric properties in the deformed scene using the relative transformation of neighboring surfels: $M_j = T^{-1}_{q}T_j$ with respect to the query surfel.

Given $M_j$, we take the linear component \( \mathbf{L}_j \in \mathbb{R}^{3 \times 3} \) of $M_j$ to analyze local deformation. The principal stretches \( \sigma_1, \sigma_2 \) are derived from the eigenvalues of the stretching matrix \( \mathbf{S} \), computed by projecting and averaging neighbor transformations in the local tangent plane:
\begin{equation}
\mathbf{S} =  \sum_{j \in \mathcal{N}} k_{j} \mathbf{P} \mathbf{L}_j \mathbf{P}^T \mathbf{P}\mathbf{L}_j^T \mathbf{P}^T,
\end{equation}
where $k_j$ is a radial basis kernel weight, \( \mathbf{P} = \mathbf{I} - \nabla f_{\text{SDF}} {\nabla f_{\text{SDF}}}^T \) is the tangent projection matrix. Similarly, principal curvatures \( \kappa_1, \kappa_2 \) are extracted from the eigenvalues of the projected Hessian:
\begin{equation}
\mathbf{H}_{\text{tangent}} = \mathbf{P}\mathbf{J} \mathbf{H} \mathbf{J}^T\mathbf{P}^T,
\end{equation}
where $\mathbf{J}$ is approximated Jacobian using $\{M_j\}_{j\in \mathcal{N}}$ and $ \mathbf{H}= \nabla^2 f_{\text{SDF}}$ .
The surfel scale is adjusted to account for both stretching and curvature. Instead of using a linear scaling term, we apply an inverse curvature-dependent scaling factor to more effectively reduce elongation in regions of high curvature:
\begin{equation}
s_{1}^{adj} = s_{1} \cdot \sigma_1 \frac{1}{1 + w_s |\kappa_1|_1}, \quad
s_{2}^{adj} = s_{2} \cdot \sigma_2 \frac{1}{1 + w_s |\kappa_2|_1},
\end{equation}
where $w_s$ denotes a scalar controlling how much the scales are adjusted.

Rotation adjustments ensure alignment with principal curvature directions while preserving the original normal. Instead of re-ranking the principal directions, we first rank the surfel basis vectors \( (\mathbf{r}_{1}, \mathbf{r}_{2}) \) based on their corresponding scale values \( (s_{1}, s_{2}) \) to obtain \((\mathbf{r}_{(1)}, \mathbf{r}_{(2)}) \) and then align them with \( (\mathbf{h}_1, \mathbf{h}_2) \) which is the principal curvature direction (\(\mathbf{H}_{\text{tangent}}\)'s eigenvectors). The misalignment axis $\mathbf{a}_i$ and angle $\theta_i$ are therefore computed as:
\begin{equation}
\mathbf{a}_i = \frac{\mathbf{r}_{(i)} \times \mathbf{h}_i}{|\mathbf{r}_{(i)} \times \mathbf{h}_i|_1}, \quad
\theta_i = \cos^{-1}(\mathbf{r}_{(i)} \cdot \mathbf{h}_i) \cdot w_q,
\end{equation}
for \( i \in \{1,2\} \), leading to the small rotation quaternions:
\begin{equation}
\hat q_i = \left( \cos \frac{\theta_i}{2}, \sin \frac{\theta_i}{2} \mathbf{a}_i \right).
\end{equation}

Since Gaussian rotation $\{\mathbf{R}\}$ is stored in quaternion form $\{\mathbf{q}\}$, the adjusted rotation is obtained via quaternion multiplication:

\begin{equation}
\mathbf{q}^{adj} = \hat q_1 \hat q_2 \mathbf{q}.
\end{equation}

Surfels without valid neighbors or bounded to the hair mesh are excluded from deformation updates. The use of inverse curvature-based scale reduction prevents excessive surfel elongation in highly curved regions, while small corrective rotations ensure alignment with principal curvature directions. This formulation enables smooth and consistent adaptation of surfel-based representations to underlying deformations while preserving local surface properties.

\subsection{Self-Supervised Training Phase}
To enhance the robustness of our avatar across diverse facial expressions, we introduce a novel self-supervised training strategy. Following Gaussian Avatar \cite{qian_gaussianavatars_2024}, we deform the Gausian with FLAME parameters but instead of relying on direct image comparisons, we apply the normal consistency and distortion regularization constraints introduced in 2DGS. These ensure that under deformation, the Gaussians remain tangential to the underlying surface. 

To ensure appearance consistency, we devise a loss function to compare pixels rendered from Gaussian rasterization and canonical volumetric rendering. Specifically, to render a sampled ray from deformed scenes, we warp the ray by performing an inverse mapping to the ray based on the transformation of the triangle where the ray intersects. To maintain efficiency, we only sample pixels from regions where the normal consistency and distortion loss are high, which often contain the most discolorations and artifacts. Meanwhile, we avoid sampling from hair regions. The appearance loss is defined by:

\begin{equation}
    \mathcal{L}_{\text{appearance}} = \sum_{\mathbf{h,w}} | \hat{C}(\gamma(\mathbf{r_{h,w}})) - \mathit{I}_{h,w} |^2,
\end{equation}
where $h,w$ are sampled pixel coordinates and $\gamma(\cdot)$ is the warping procedure and $\mathit{I}$ is the rasterized image. The overall loss function for the second training phase is defined as:

\begin{equation}
    \mathcal{L}_{total} = \lambda_n \mathcal{L}_{\text{n}} + \lambda_d \mathcal{L}_{\text{d}} + \lambda_a \mathcal{L}_{\text{appearance}}.
\end{equation}

During the second training phase, we freeze the parameters in the $f_{SDF}$ as well as the hash feature grid $\{\Phi ^l\}^L_{l=1}$. We utilize the same expression sequence from the NeRSemble dataset \cite{kirschstein_nersemble_2023}, though alternative datasets such as FaMoS \cite{bolkart_instant_2023} and CoMA \cite{ranjan_generating_2018} can also be incorporated to add variation in training data.

\subsection{Adaptability}

Our pipeline's core strength is its strong adaptability to generalize across common reconstruction scenarios while markedly improving data efficiency and alleviating the typical reliance on expression-rich training data.

\noindent\textbf{Multiview Reconstruction.}
In the standard multiview setting, our method directly follows conventional Gaussian Splatting and static SDF reconstruction protocols, and no additional modifications are required.

\noindent\textbf{Monocular Reconstruction.}
In contrast to existing monocular approaches that depend on long expression sequences to resolve expression dynamics, our method only requires moderate head rotations for full coverage of the target head, substantially reducing data demands. As premise, our pipeline requires a backward-warping mechanism for accurately reconstructing SDF through dynamic ray rendering. We integrate the pretrained deformation prior from MonoNPHM~\cite{giebenhain_mononphm_2024} to backward-warp rays into canonical space, enabling the recovery of a neutral SDF under pose variation.  We follow GaussianAvatars~\cite{qian_gaussianavatars_2024} for dynamic Gaussian optimization and apply a super-resolution stage to counteract motion blur and enhance fine details, maintaining reconstruction quality despite limited observations.

\noindent\textbf{One‐shot Reconstruction.}
For one-shot generation of animatable avatars, methods such as \cite{schoneveld2025sheapselfsupervisedheadgeometry, he_lam_2025, saunders_gasp_2024} require training on expression-rich data, which often leads to geometric inaccuracies. In contrast, approaches like \cite{lyu_facelift_2025, kirschstein_gghead_2024} can produce geometrically accurate avatars but lack animatability. Our pipeline bridges this gap by transforming static avatar generations into fully animatable Gaussian representations.

As a demonstration, we integrate the GenHead module from Portrait-4D~\cite{deng_portrait4d_2024} to synthesize pseudo multi-view images using a triplane head representation conditioned on FLAME parameters. A lightweight fine-tuning on the NeRSemble dataset mitigates the domain gap and improves geometric consistency across synthesized views. Since GenHead operates at a relatively low spatial resolution, we further incorporate a super-resolution stage to recover high-frequency appearance details without requiring additional input images. This design enables a stable and reliable reconstruction pipeline even under extreme data scarcity.

With these adaptations, our method supports reliable reconstruction under multiview, monocular, and one-shot scenarios, consistently improving data efficiency while reducing the dependence on explicit expression supervision or densely captured expression sequences.
\section{Experiments}
\begin{figure*}[t]
    \centering
    \begin{subfigure}[b]{0.13\linewidth}
        \includegraphics[width=\linewidth, clip,trim=5 190 40 20]{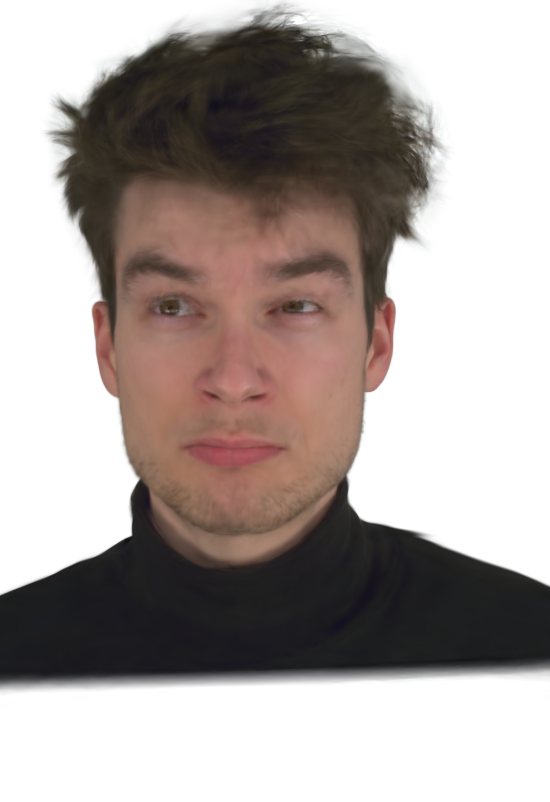}
    \end{subfigure}\hspace{0.5mm}
    \begin{subfigure}[b]{0.13\linewidth}
        \includegraphics[width=\linewidth, clip,trim=5 190 40 20]{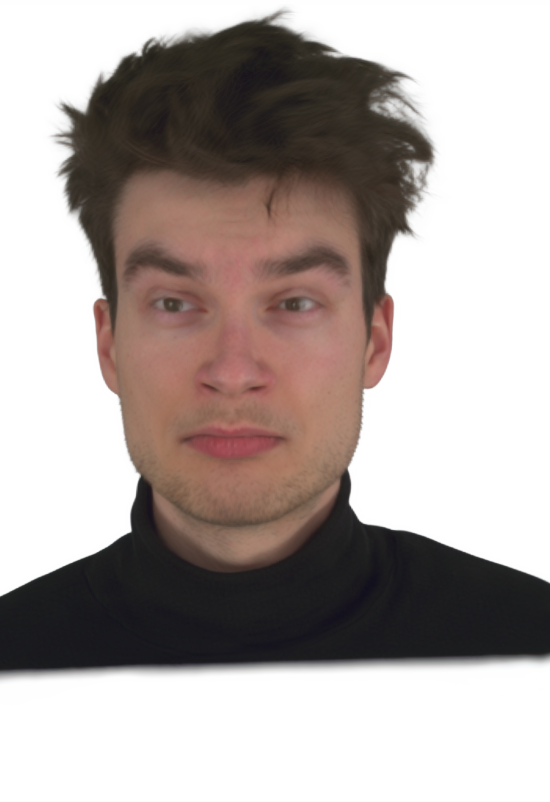}
    \end{subfigure}\hspace{0.5mm}
    \begin{subfigure}[b]{0.13\linewidth}
        \includegraphics[width=\linewidth, clip,trim=5 210 30 0]{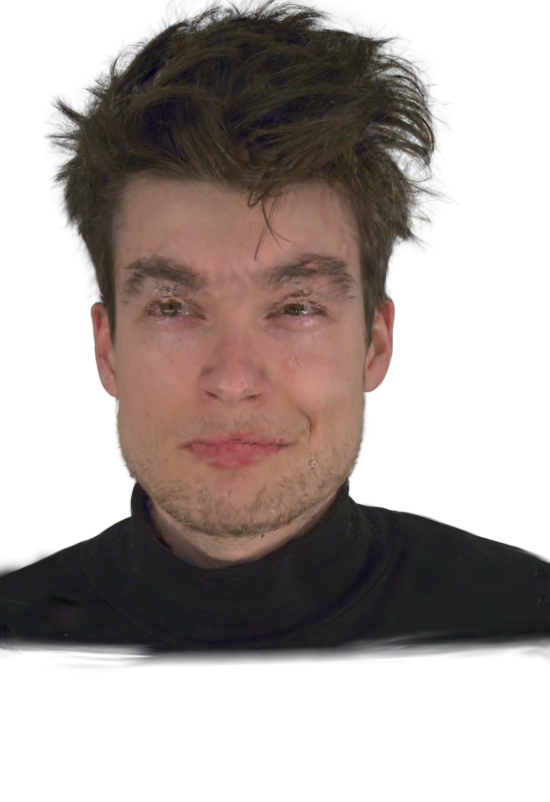}
    \end{subfigure}\hspace{0.5mm}
    \begin{subfigure}[b]{0.13\linewidth}
        \includegraphics[width=\linewidth, clip,trim=5 190 40 20]{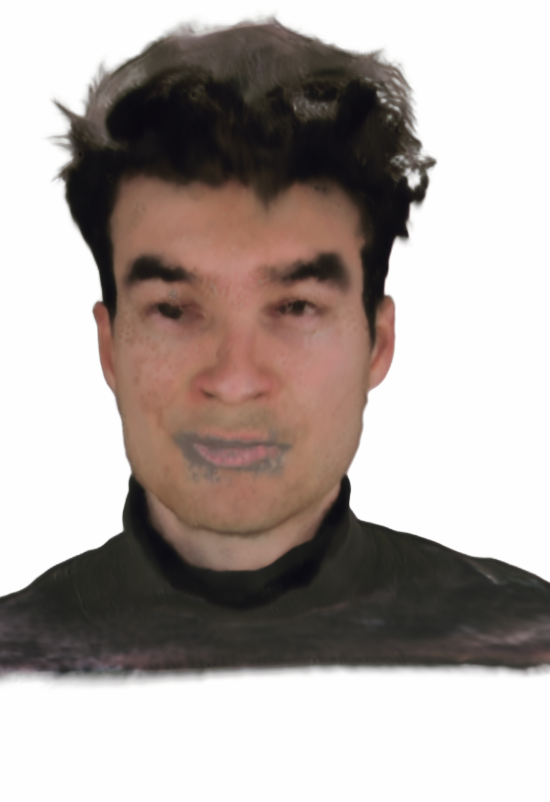}
    \end{subfigure}\hspace{0.5mm}
    \begin{subfigure}[b]{0.13\linewidth}
        \includegraphics[width=\linewidth, clip,trim=5 190 50 20]{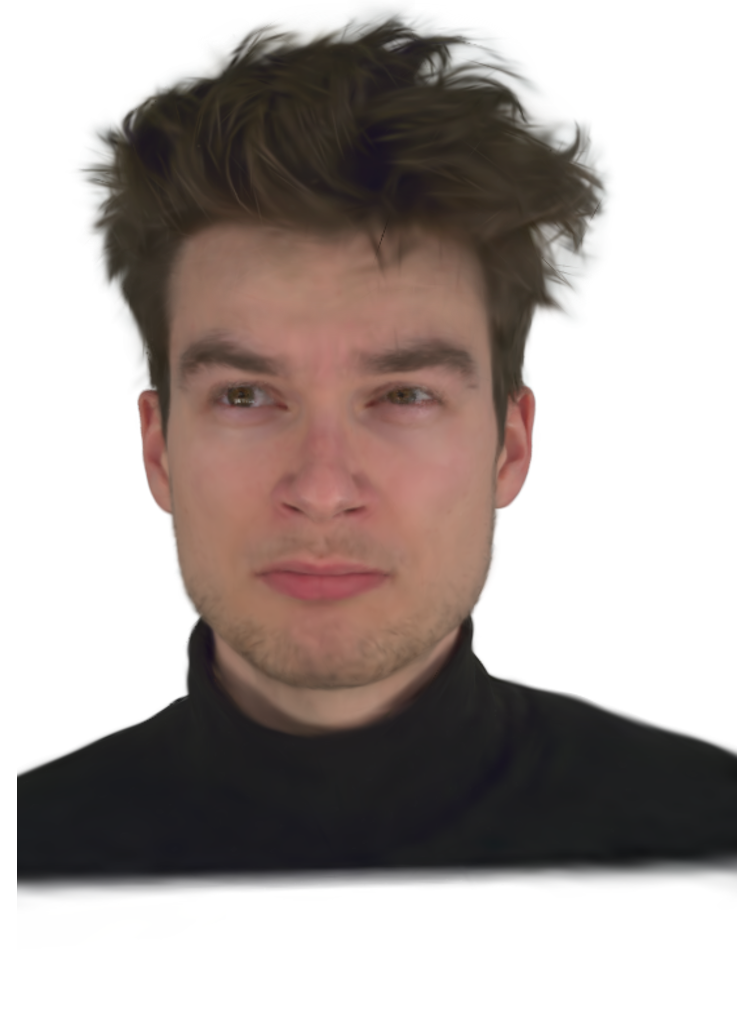}
    \end{subfigure}\hspace{0.5mm} 
    \begin{subfigure}[b]{0.13\linewidth}
        \includegraphics[width=\linewidth, clip,trim=5 190 40 20]{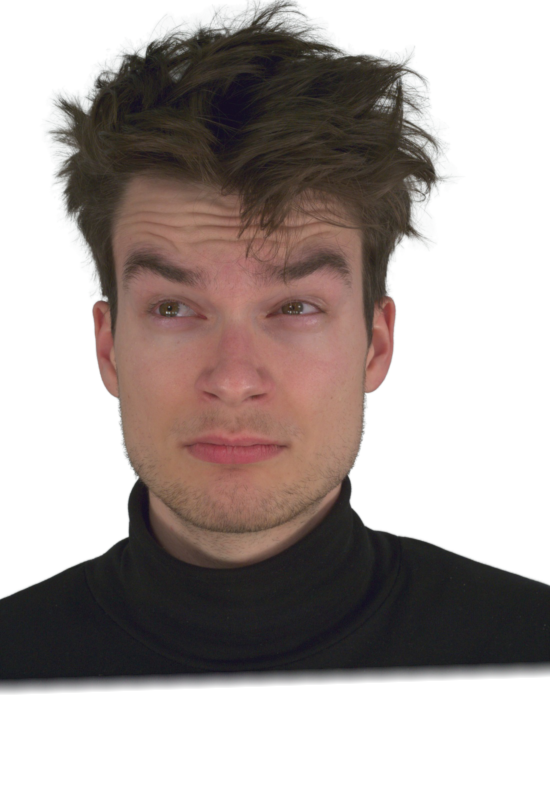}
    \end{subfigure}

    \begin{subfigure}[b]{0.13\linewidth}
        \includegraphics[width=\linewidth, clip,trim=60 200 70 60]{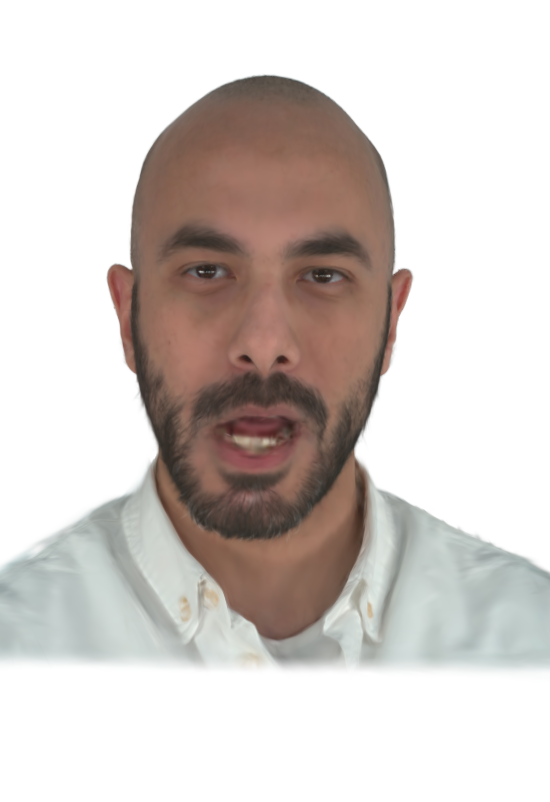}
    \end{subfigure}\hspace{0.5mm}
    \begin{subfigure}[b]{0.13\linewidth}
        \includegraphics[width=\linewidth, clip,trim=60 200 70 60]{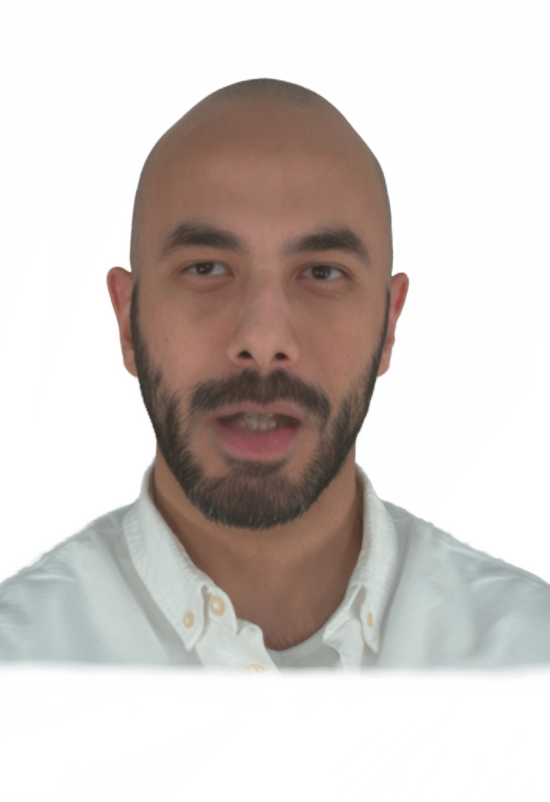}
    \end{subfigure}\hspace{0.5mm}
    \begin{subfigure}[b]{0.13\linewidth}
        \includegraphics[width=\linewidth, clip,trim=60 200 70 60]{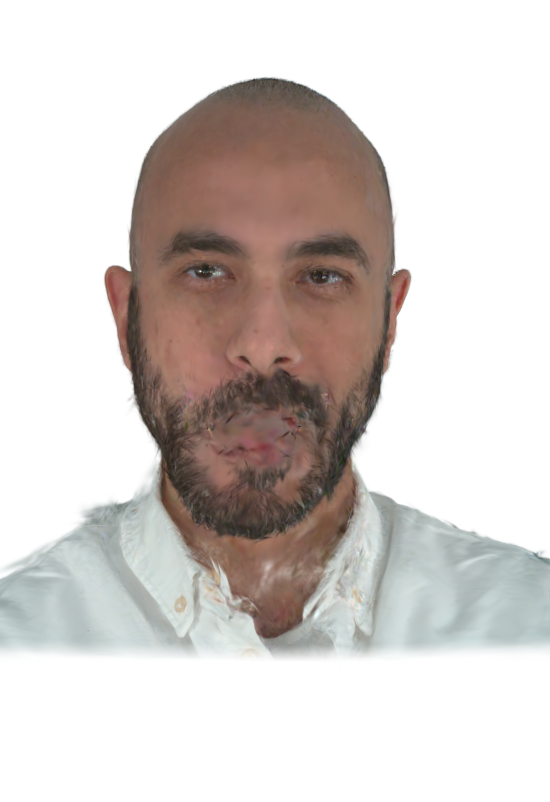}
    \end{subfigure}\hspace{0.5mm}
    \begin{subfigure}[b]{0.13\linewidth}
        \includegraphics[width=\linewidth, clip,trim=60 200 70 60]{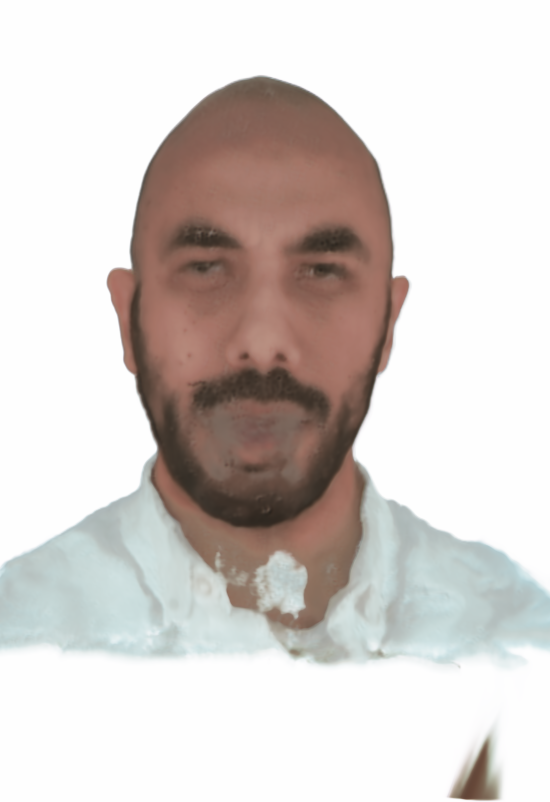}
    \end{subfigure}\hspace{0.5mm}
    \begin{subfigure}[b]{0.13\linewidth}
        \includegraphics[width=\linewidth, clip,trim=60 200 70 60]{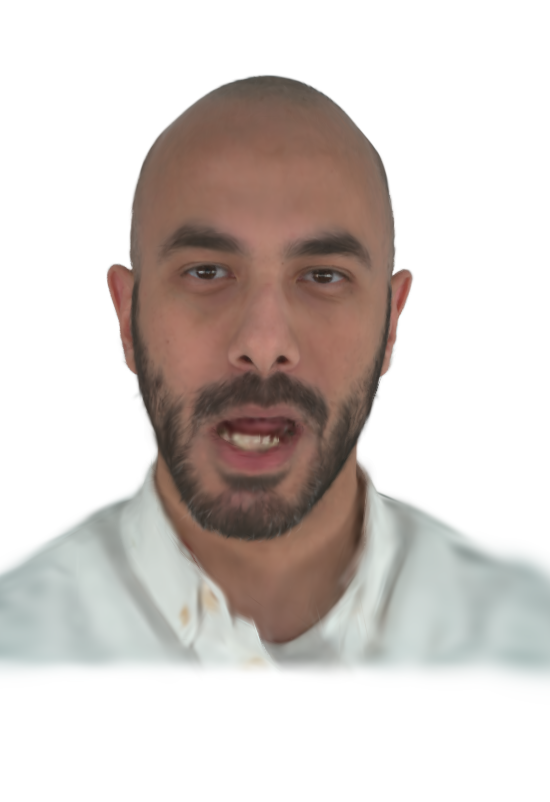}
    \end{subfigure}\hspace{0.5mm}
    \begin{subfigure}[b]{0.13\linewidth}
        \includegraphics[width=\linewidth, clip,trim=60 200 70 60]{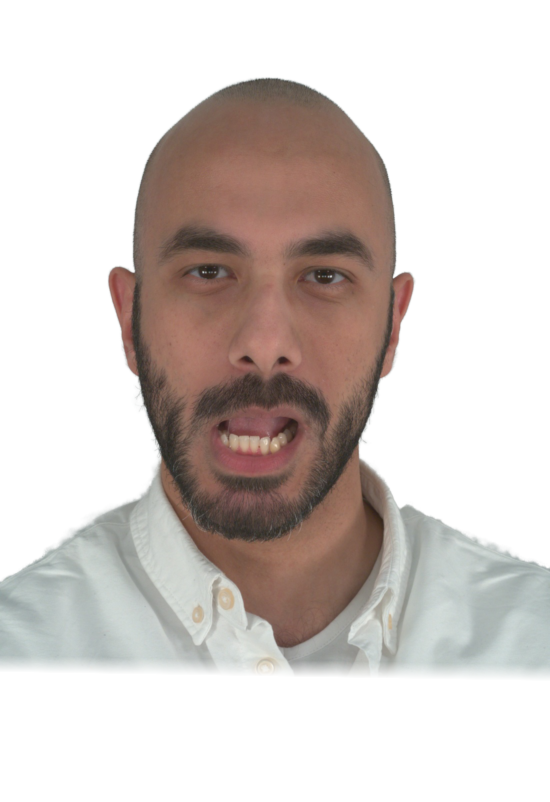}
    \end{subfigure}


    \begin{subfigure}[b]{0.13\linewidth}
        \includegraphics[width=\linewidth,height=2.55cm, clip,trim=10 120 10 120]{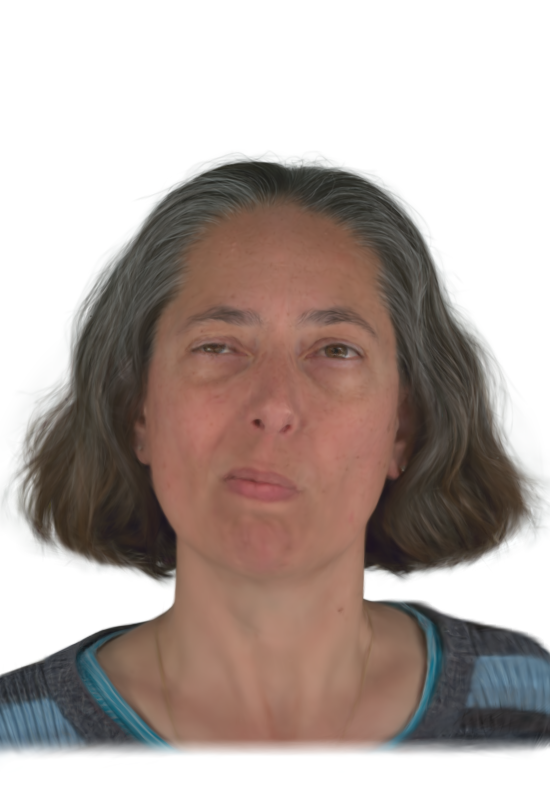}
    \end{subfigure}\hspace{0.5mm}
    \begin{subfigure}[b]{0.13\linewidth}
        \includegraphics[width=\linewidth, height=2.55cm, clip,trim=10 120 10 120]{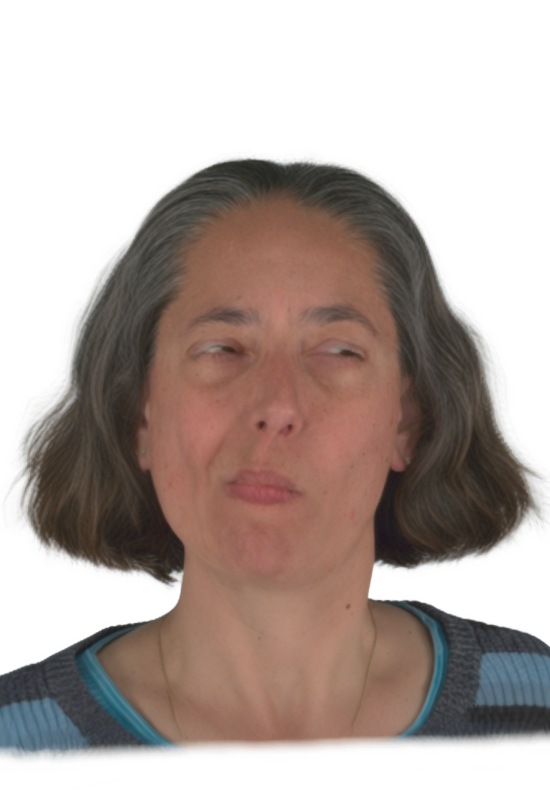}
    \end{subfigure}\hspace{0.5mm}
    \begin{subfigure}[b]{0.13\linewidth}
        \includegraphics[width=\linewidth, height=2.55cm, clip,trim=10 150 10 120]{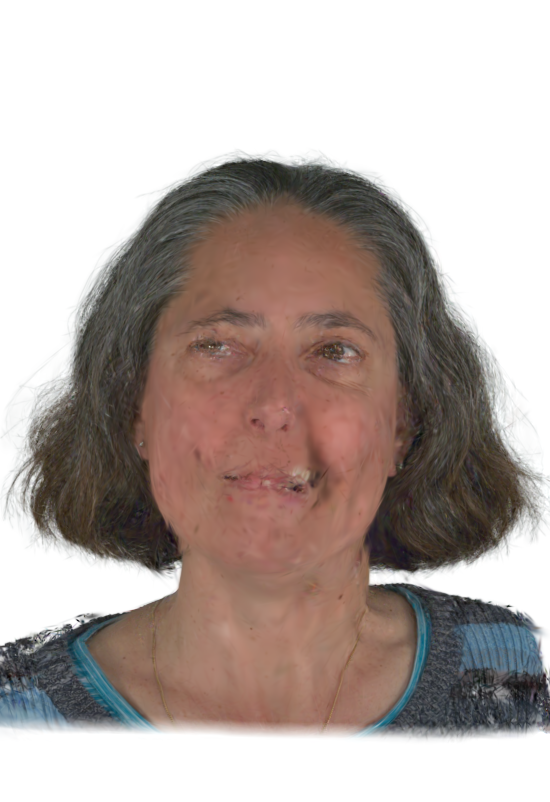}
    \end{subfigure}\hspace{0.5mm}
    \begin{subfigure}[b]{0.13\linewidth}
        \includegraphics[width=\linewidth, height=2.55cm, clip,trim=10 120 10 120]{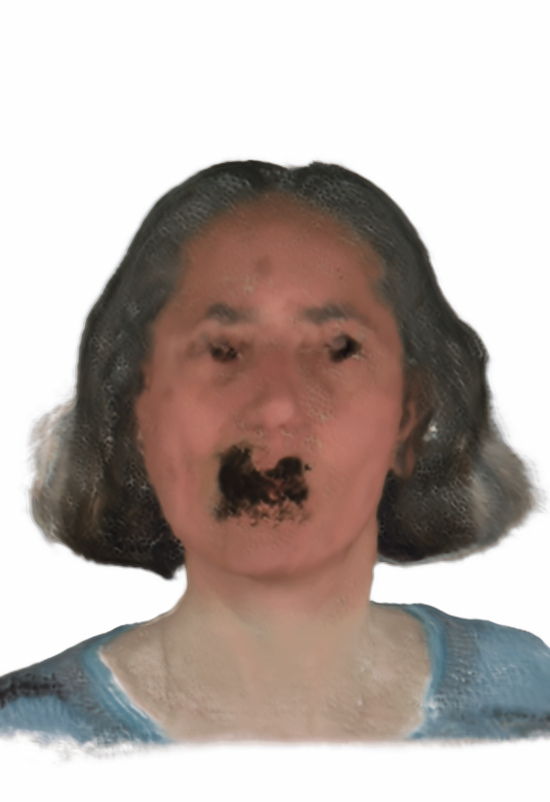}
    \end{subfigure}\hspace{0.5mm}
    \begin{subfigure}[b]{0.13\linewidth}
        \includegraphics[width=\linewidth, height=2.55cm, clip,trim=10 120 10 120]{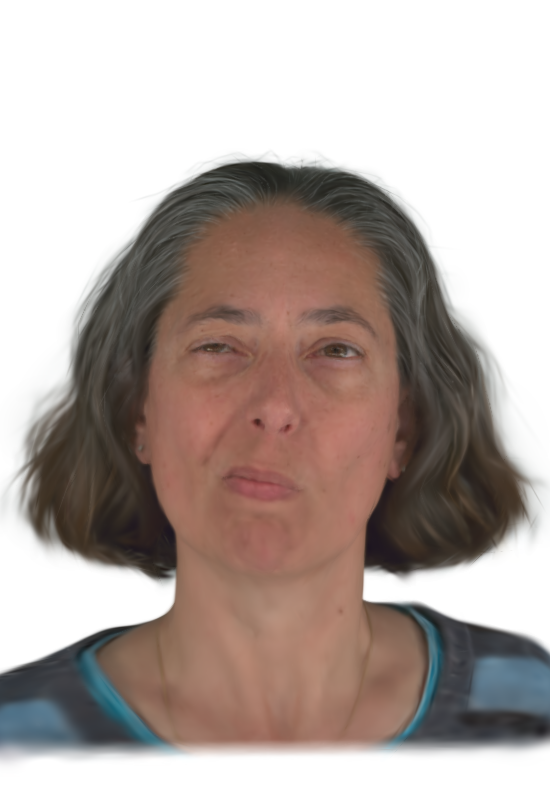}
    \end{subfigure}\hspace{0.5mm}
    \begin{subfigure}[b]{0.13\linewidth}
        \includegraphics[width=\linewidth, height=2.55cm, clip,trim=10 120 10 120]{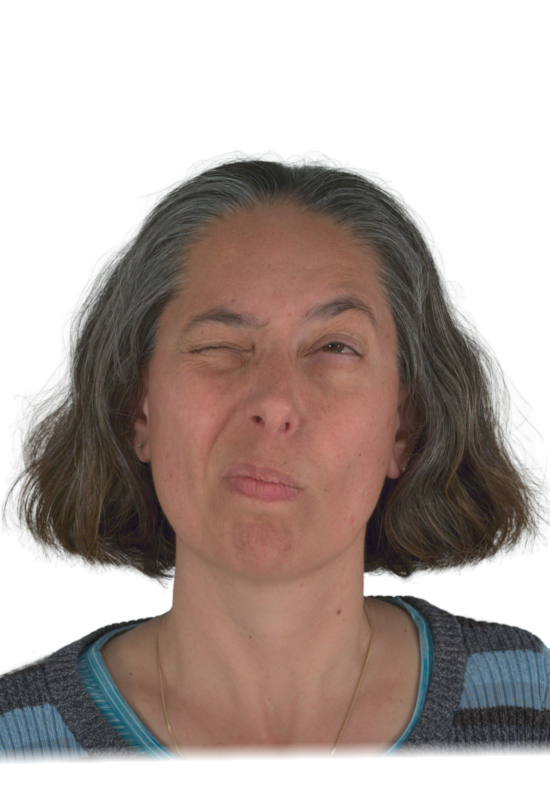}
    \end{subfigure}

    \scriptsize
    \par\smallskip
    \parbox{0.13\linewidth}{\centering GA}
    \hspace{0.5mm}
    \parbox{0.13\linewidth}{\centering GHA}
    \hspace{0.5mm}
    \parbox{0.13\linewidth}{\centering GA 1 frame}
    \hspace{0.5mm}
    \parbox{0.13\linewidth}{\centering GHA 1 frame}
    \hspace{0.5mm}
    \parbox{0.13\linewidth}{\centering \textbf{Ours\textsubscript{SF}}}
    \hspace{0.5mm}
    \parbox{0.13\linewidth}{\centering GT}
    
    \caption{Qualitative comparisons for self-reenactment quality under Multiview input setting.}
    \label{fig:self-re}
\end{figure*}

\begin{figure*}[t]
    \centering

     \begin{subfigure}[b]{0.13\linewidth}
        \includegraphics[width=\linewidth, clip,trim=5 200 40 20]{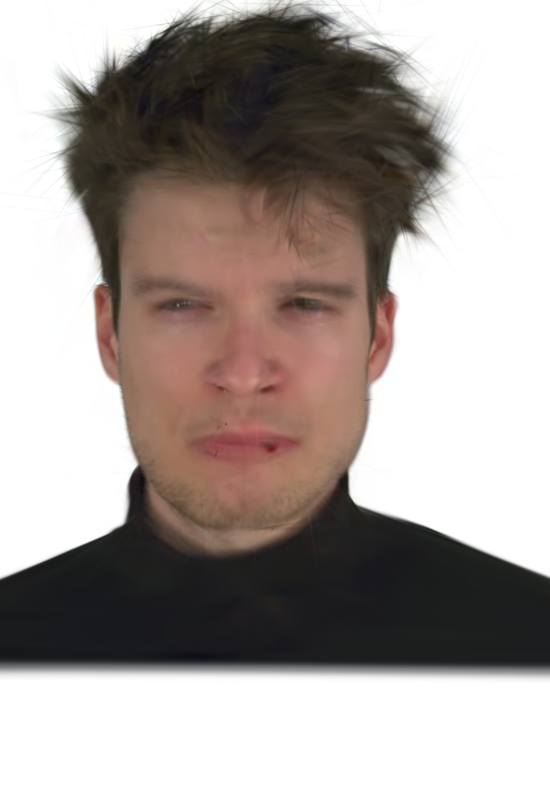}
    \end{subfigure}\hspace{0.5mm}
    \begin{subfigure}[b]{0.13\linewidth}
        \includegraphics[width=\linewidth, clip,trim=0 190 40 0]{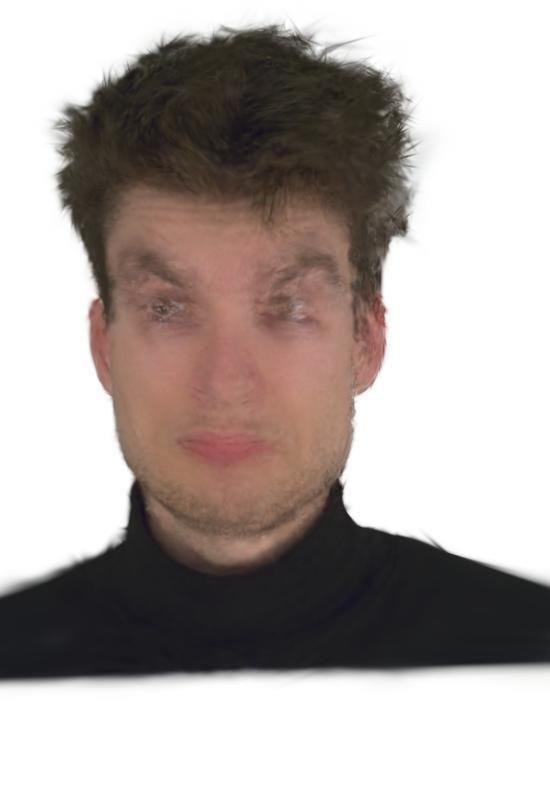}
    \end{subfigure}\hspace{0.5mm}
    \begin{subfigure}[b]{0.13\linewidth}
        \includegraphics[width=\linewidth, clip,trim=0 190 40 0]{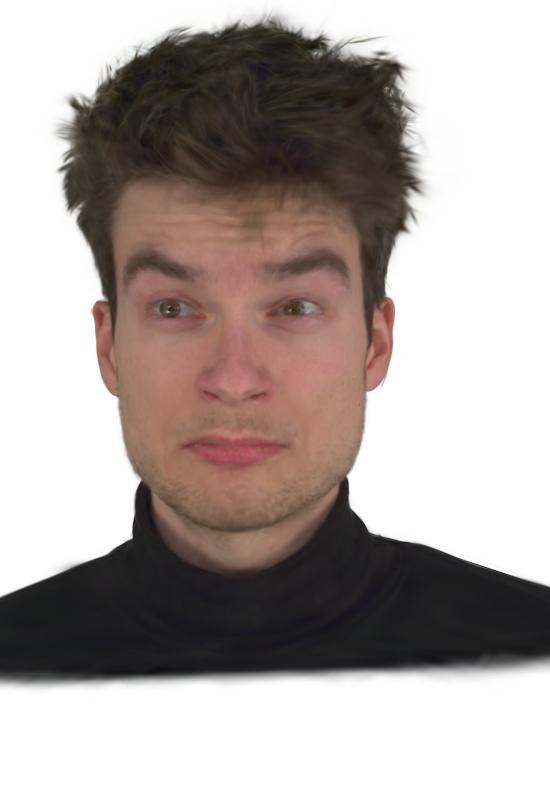}
    \end{subfigure}\hspace{0.5mm}
    \begin{subfigure}[b]{0.13\linewidth}
        \includegraphics[width=\linewidth, clip,trim=0 190 40 0]{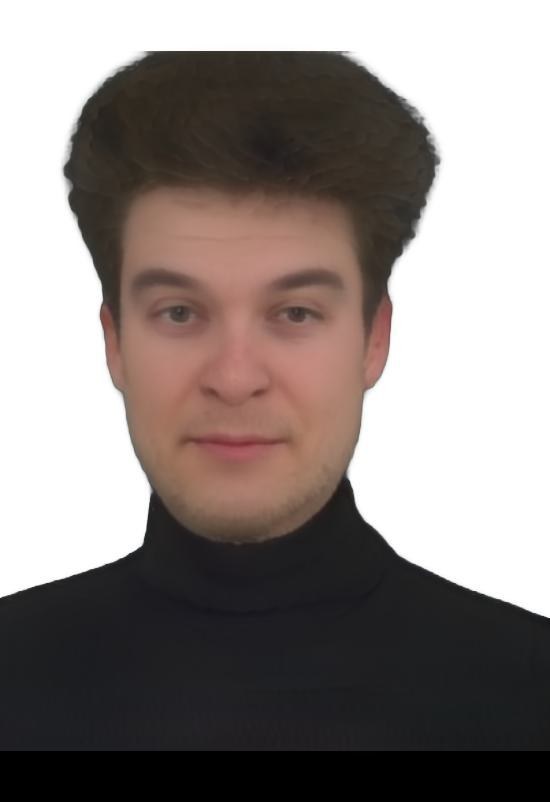}
    \end{subfigure}\hspace{0.5mm}
    \begin{subfigure}[b]{0.13\linewidth}
        \includegraphics[width=\linewidth, clip,trim=5 190 40 20]{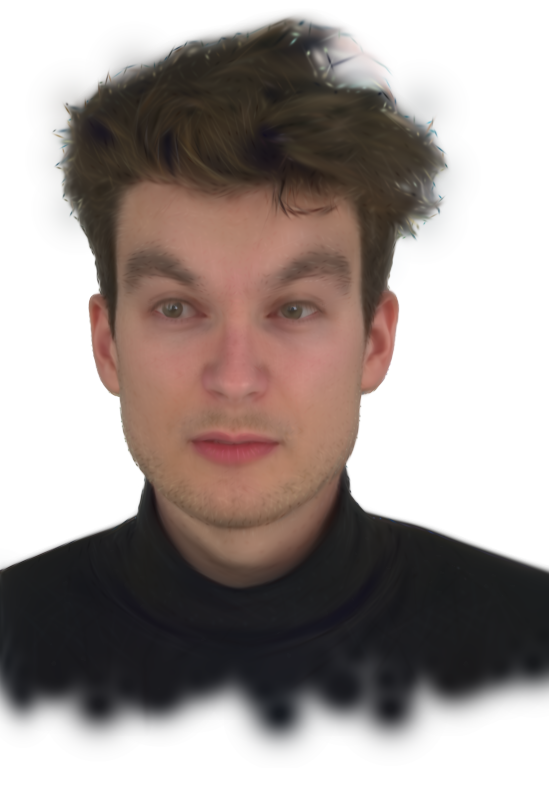}
    \end{subfigure}\hspace{0.5mm}
    \begin{subfigure}[b]{0.13\linewidth}
        \includegraphics[width=\linewidth, clip,trim=5 210 0 5]{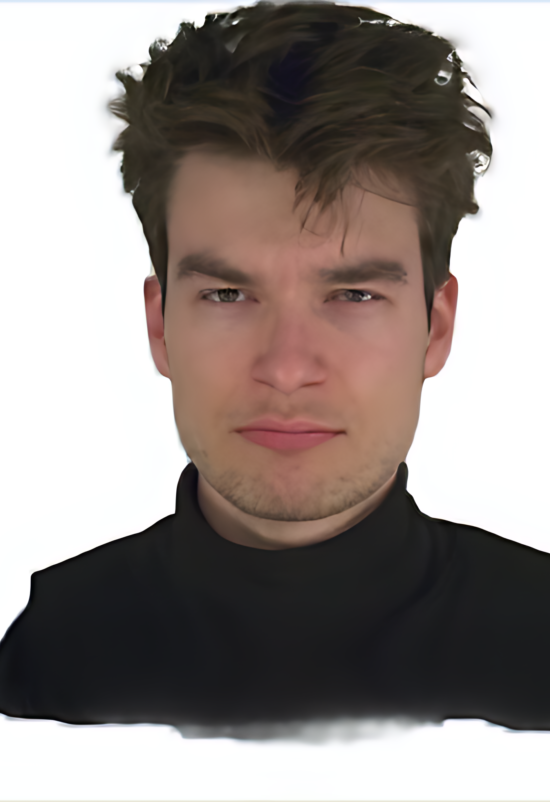}
    \end{subfigure}\hspace{0.5mm} 
    \begin{subfigure}[b]{0.13\linewidth}
        \includegraphics[width=\linewidth, clip,trim=5 190 40 20]{results/104gt_00125.png}
    \end{subfigure}
    \begin{subfigure}[b]{0.13\linewidth}
        \includegraphics[width=\linewidth, clip,trim=60 210 20 40]{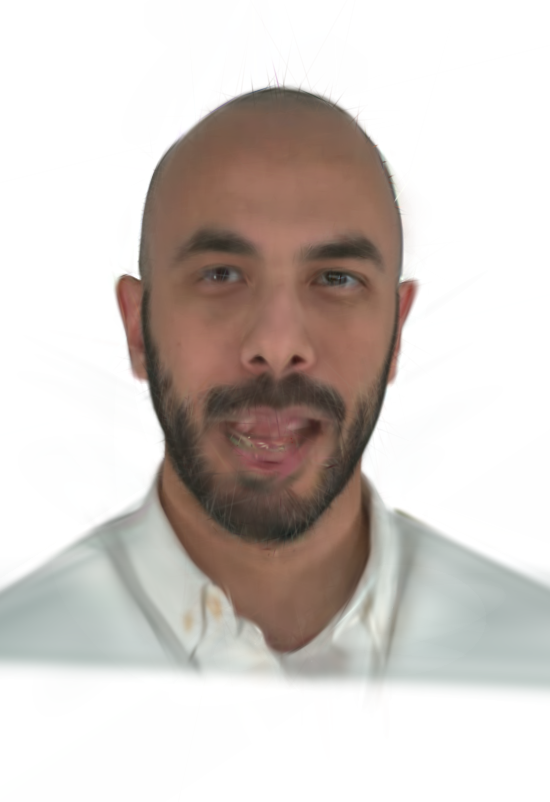}
    \end{subfigure}\hspace{0.5mm}
    \begin{subfigure}[b]{0.13\linewidth}
        \includegraphics[width=\linewidth, clip,trim=60 210 20 40]{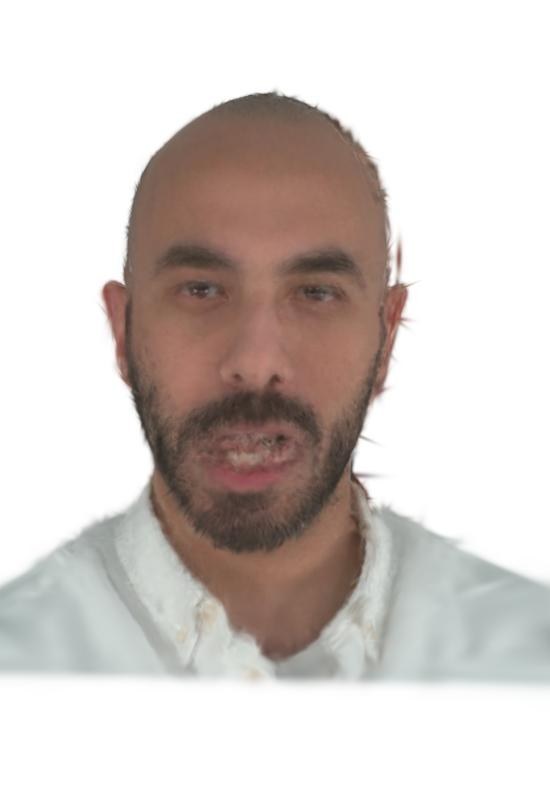}
    \end{subfigure}\hspace{0.5mm}
    \begin{subfigure}[b]{0.13\linewidth}
        \includegraphics[width=\linewidth, clip,trim=60 210 20 40]{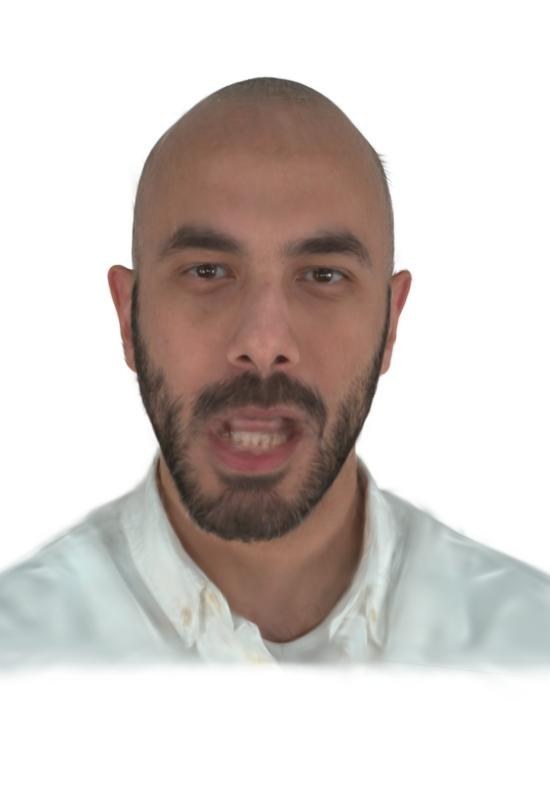}
    \end{subfigure}\hspace{0.5mm}
    \begin{subfigure}[b]{0.13\linewidth}
        \includegraphics[width=\linewidth, clip,trim=50 230 50 50]{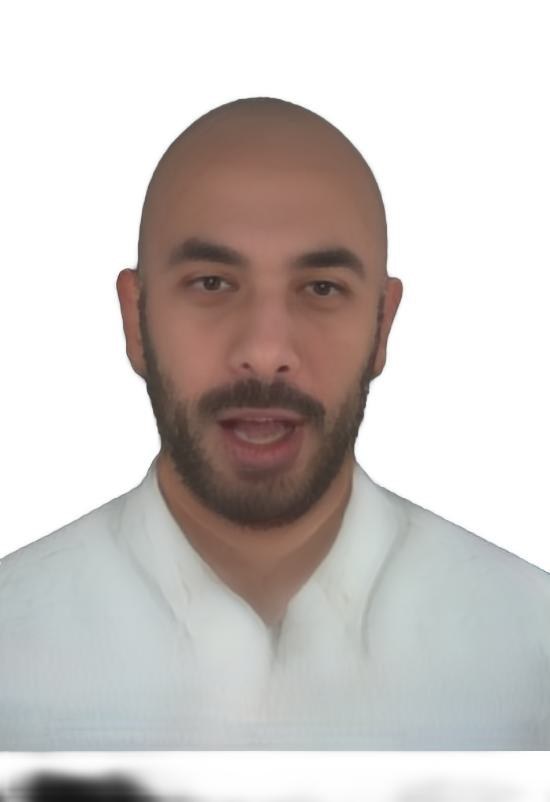}
    \end{subfigure}\hspace{0.5mm}
    \begin{subfigure}[b]{0.13\linewidth}
        \includegraphics[width=\linewidth, clip,trim=60 210 20 40]{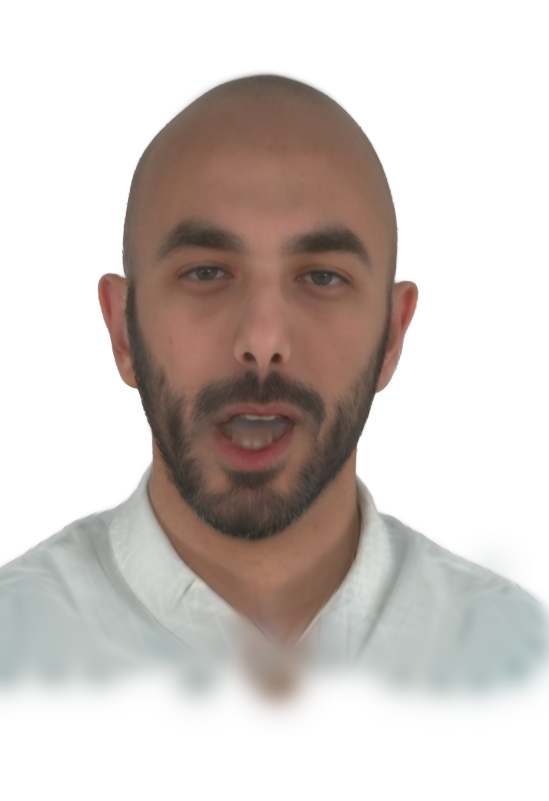}
    \end{subfigure}\hspace{0.5mm}
    \begin{subfigure}[b]{0.13\linewidth}
        \includegraphics[width=\linewidth, clip,trim=30 190 0 20]{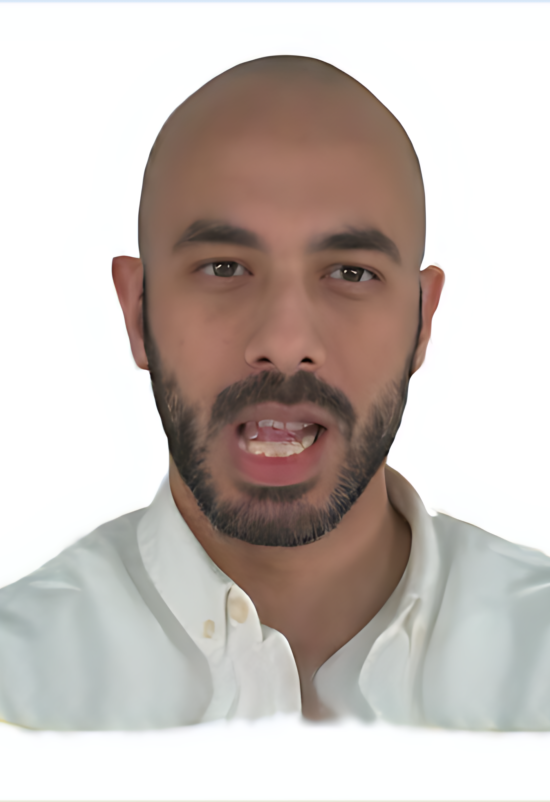}
    \end{subfigure}\hspace{0.5mm}
    \begin{subfigure}[b]{0.13\linewidth}
        \includegraphics[width=\linewidth, clip,trim=60 210 20 40]{results/218gt_00066.png}
    \end{subfigure}

    \begin{subfigure}[b]{0.13\linewidth}
        \includegraphics[width=\linewidth, clip,trim=10 150 10 120]{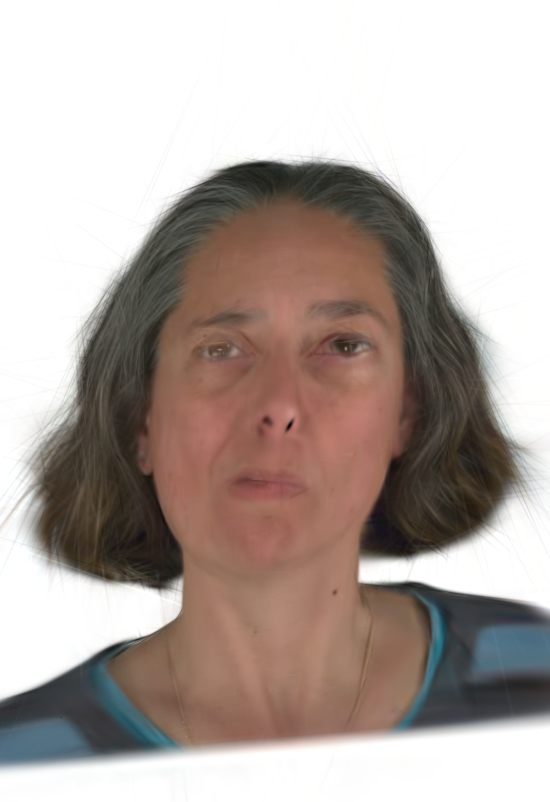}
    \end{subfigure}\hspace{0.5mm}
    \begin{subfigure}[b]{0.13\linewidth}
        \includegraphics[width=\linewidth, clip,trim=10 160 10 80]{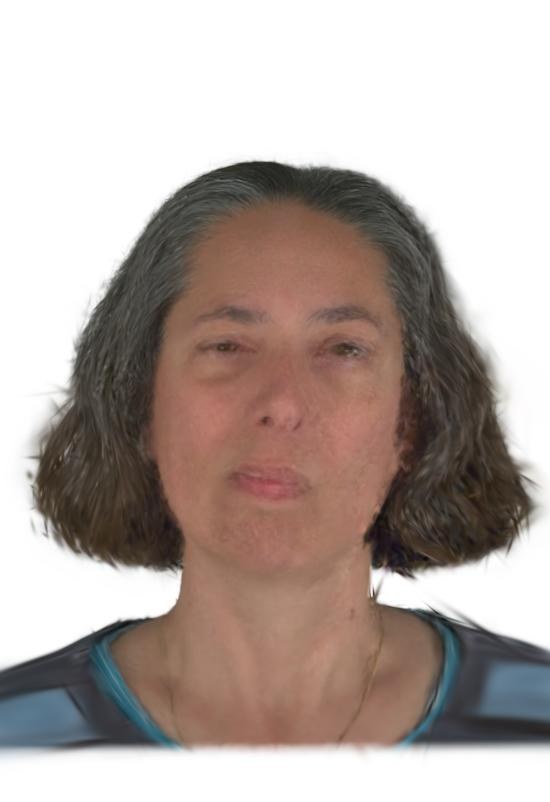}
    \end{subfigure}\hspace{0.5mm}
    \begin{subfigure}[b]{0.13\linewidth}
        \includegraphics[width=\linewidth, clip,trim=10 160 10 80]{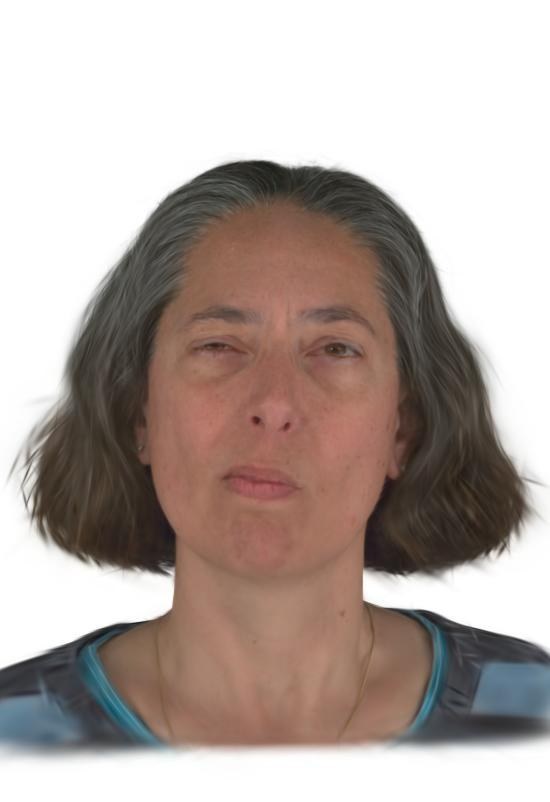}
    \end{subfigure}\hspace{0.5mm}
    \begin{subfigure}[b]{0.13\linewidth}
        \includegraphics[width=\linewidth, clip,trim=10 150 10 120]{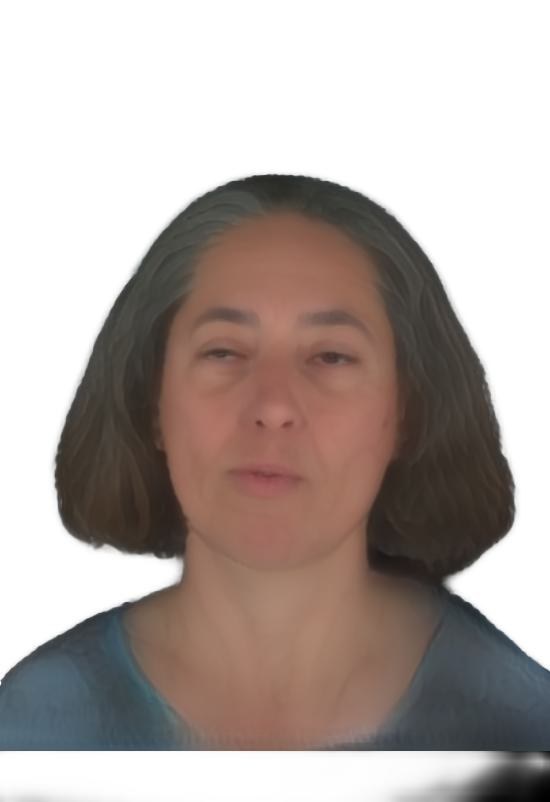}
    \end{subfigure}\hspace{0.5mm}
    \begin{subfigure}[b]{0.13\linewidth}
        \includegraphics[width=\linewidth, clip,trim=10 150 10 120]{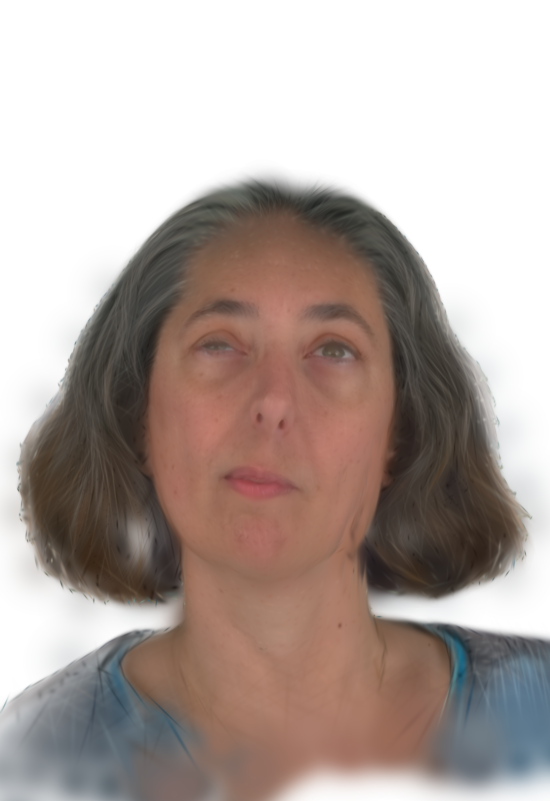}
    \end{subfigure}\hspace{0.5mm}
    \begin{subfigure}[b]{0.13\linewidth}
        \includegraphics[width=\linewidth, clip,trim=10 190 10 80]{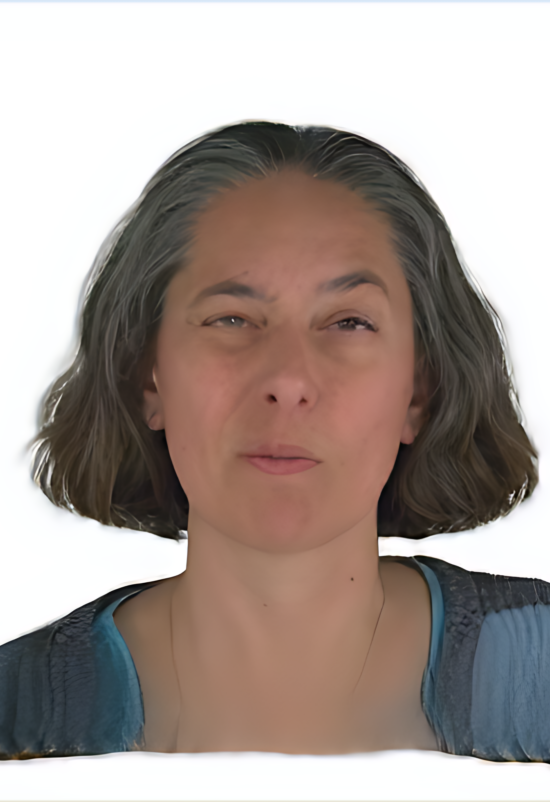}
    \end{subfigure}\hspace{0.5mm}
    \begin{subfigure}[b]{0.13\linewidth}
        \includegraphics[width=\linewidth, clip,trim=10 150 10 120]{results/304gt_00085.png}
    \end{subfigure}
    
    \scriptsize
    \par\smallskip
    \parbox{0.13\linewidth}{\centering (a) FlashAvatar}
    \hspace{0.5mm}
    \parbox{0.13\linewidth}{\centering (a) RGBAvatar}
    \hspace{0.5mm}
    \parbox{0.13\linewidth}{\centering (a) \textbf{Ours\textsubscript{Mono}}}
    \hspace{0.5mm}
    \parbox{0.13\linewidth}{\centering (b) GPAvatar}
    \hspace{0.5mm}
    \parbox{0.13\linewidth}{\centering (b) LAM}
    \hspace{0.5mm}
    \parbox{0.13\linewidth}{\centering (b) \textbf{Ours\textsubscript{one-shot}}}
    \hspace{0.5mm}
    \parbox{0.13\linewidth}{\centering GT}
    
    \caption{Qualitative comparisons for self-reenactment quality under different input types. Results (a) refers to monocular reconstructure setting, results (b) refers to one-shot reconstruction setting.}
    \label{fig:mono_one}
\end{figure*}

\begin{table*}[t]
\centering
\small
\begin{tabular}{cc ccc ccc }
\hline
\multirow{2}{*}{\centering Method} & \multirow{2}{*}{\centering Input Type} &
\multicolumn{3}{c}{Novel Expression (frontal)} & \multicolumn{3}{c}{Novel Expression (all view)}  \\
&  & PSNR↑ & SSIM↑ & LPIPS↓ & PSNR↑ & SSIM↑ & LPIPS↓ \\
\hline
GA \cite{qian_gaussianavatars_2024} & MV-MF &25.94 &0.922 & 0.079 &25.32 &0.903 & 0.084 \\
GHA \cite{xu_gaussian_2024} & MV-MF & 26.21 & 0.915 & 0.076 & 24.98& 0.894 & 0.087 \\
\hline
GA \cite{qian_gaussianavatars_2024} & MV-SF & 17.75 & 0.825 & 0.200  &17.23 & 0.823 & 0.215 \\
GHA  \cite{xu_gaussian_2024} & MV-SF & 16.21 & 0.757 & 0.214 &15.29 & 0.762 & 0.243 \\
Ours & MV-SF & \textbf{25.41} & \textbf{0.896} & \textbf{0.085} & \textbf{24.27} &  \textbf{0.879} & \textbf{0.094} \\
\hline
GPAvatar \cite{chu_gpavatar_2024} & One-shot & 17.89 & 0.816 & 0.180 & 15.47 & 0.755 & 0.286 \\
LAM \cite{he_lam_2025} & One-shot & 20.73 & 0.815 & 0.154 & 15.23 & 0.732 & 0.290  \\
Ours  & One-shot & \textbf{21.04} & \textbf{0.821} & \textbf{0.125} & \textbf{18.57}& \textbf{0.785} & \textbf{0.169}\\
\hline
FlashAvatar \cite{xiang_flashavatar_2024} & SV-MF & 21.24 & 0.866 & 0.143 & 16.95 & 0.810& 0.239\\
RGBAvatar \cite{li_rgbavatar_2025} & SV-MF & 21.32 & 0.845 & 0.135 & 15.95 & 0.809 & 0.252\\
\hline
FlashAvatar \cite{xiang_flashavatar_2024} & SV-HR & 18.36 & 0.830 & 0.159 & 15.89 & 0.783& 0.271\\
RGBAvatar \cite{li_rgbavatar_2025} & SV-HR & 19.79 & 0.827 & 0.147 & 16.95 & 0.820 & 0.260\\
Ours & SV-HR & \textbf{23.90} & \textbf{0.90} & \textbf{0.089} & \textbf{22.76} & \textbf{0.881} & \textbf{0.102}\\
\hline
\end{tabular}
\caption{Quantitative comparison with state of the art on Self-Reenactment across different reconstruction settings. MV-MF: Multiview / Multi-frame, MV-SF: Multiview / Single-frame, SV-MF: Single-view / Multi-frame, SV-HR: Single-view/ head rotation data only.}
\label{tab:main_tb}
\end{table*}

\begin{figure*}[t]
    \centering
    \includegraphics[width=0.8\linewidth]{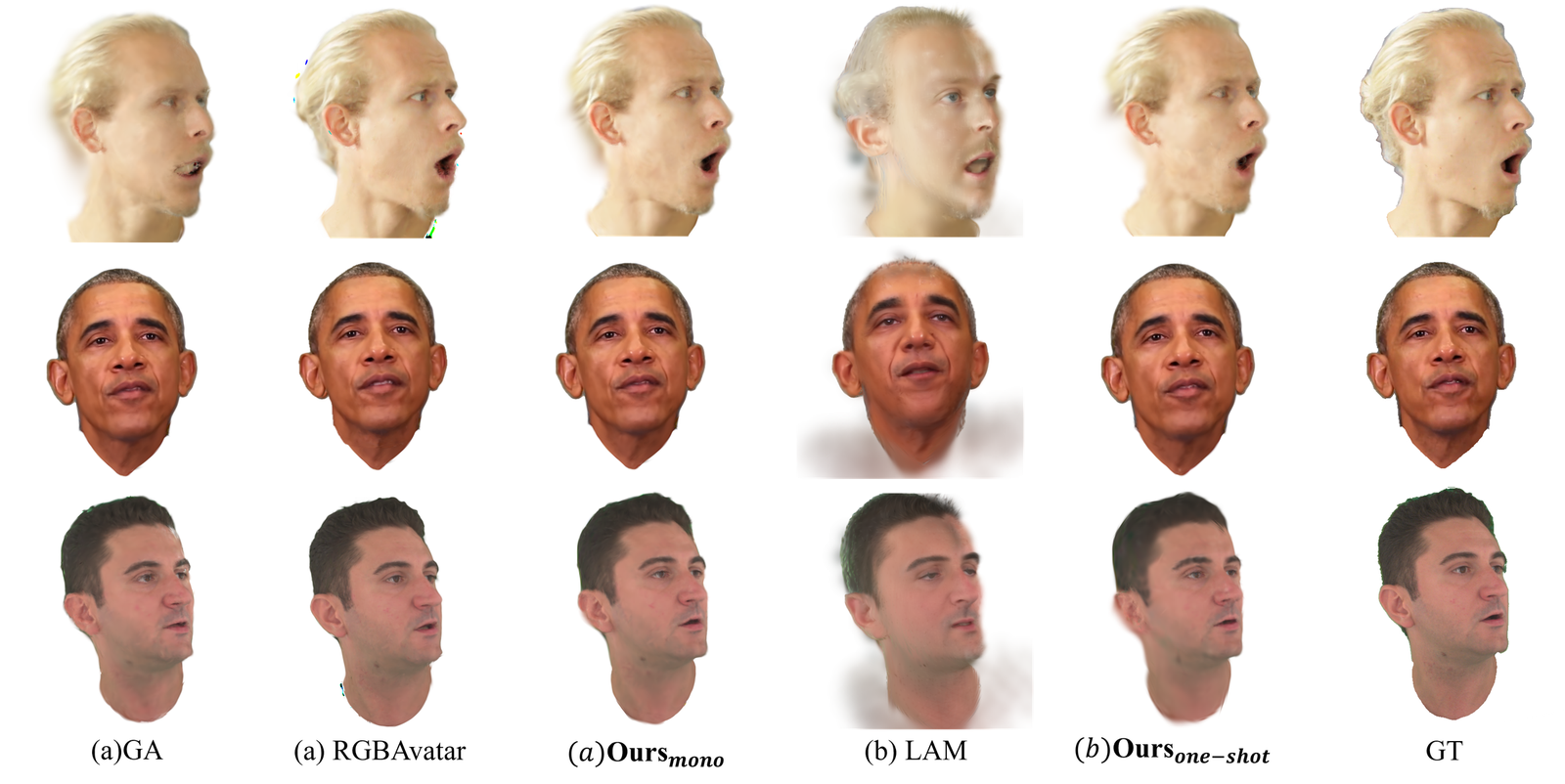}
    \caption{We also conduct qualitative assessment of our pipeline's self-reenactment performance on INSTA's data which is a less calibrated setting.}
    \label{fig:insta_d}
\end{figure*}

\begin{table}[t]
\centering
\small
\setlength{\tabcolsep}{6pt}
\begin{tabular}{cc ccc}
\toprule
Method & Input Type & PSNR$\uparrow$ & SSIM$\uparrow$ & LPIPS$\downarrow$ \\
\midrule
GA    &SV-MF  & 28.11 & 0.944 & 0.065 \\
RGBAvatar &SV-MF & 29.02 & 0.933 & 0.087 \\
Ours &SV-MF & 29.03 & 0.925 & 0.082 \\
LAM &One-shot & 22.54 & 0.839 & 0.101 \\
Ours &One-shot & 24.80 & 0.887 & 0.088  \\
\bottomrule
\end{tabular}
\caption{Quantitative comparison of reconstruction quality of INSTA data.}
\label{tab:insta}
\end{table}

\subsection{Implementation details}
In the experiments, we use 14 identities from the NeRSemble dataset, which include 11 expression sequences captured by 16 cameras covering frontal and side views of the subject. We use the preprocessed version of NeRSemble performed by \cite{qian_gaussianavatars_2024}, which selected 10 sets of expression sequences. The image is downsampled to a resolution of 550x802, and FLAME tracking is performed using VHAP \cite{qian2024vhap}. For experiments, one of the sequences is chosen as the held-out sequence for testing the self-reenactment task.

We select the NeRSemble dataset as our primary experimental benchmark because its wide range of viewpoints allows for a thorough assessment of multiview reconstruction quality across competing methods, highlighting the strengths of our approach. We further evaluate our method on the INSTA dataset \cite{zielonka_instant_2023}, which presents more challenging capture conditions and suboptimal calibration settings.

As we have an optimized FLAME coarse mesh, we pre-train the SDF on this coarse mesh and initialize the Gaussian optimization process with a pre-trained SDF. This guarantees successful convergence and allows the SDF to provide more accurate guidance to 2DGS in earlier stages of the optimization process, and can be continuously fine-tuned. We disabled 3DGS's densification strategy but instead initialize 150,000 points, and randomly assign them to different mesh triangles based on their scale, also meshes labeled as face and closer to the camera get a higher number of initialized points. During optimization, we allow Gaussians to rebind to different triangles based on distance. This is because we found that newly added Gaussians are harder to optimize, leading to lower reconstruction quality. Since the mouth interior is unobserved by our pipeline, we add a new mesh inside the mouth and rig it to the lower jaw. We pretrain a mouth interior Gaussian prior, which is added to our avatar during rendering. Finally, we darken Gaussians at areas with high distortion to mimic wrinkle effect as high curvature often associates with these areas, indicating possible skin shading. Further details can be found in supplementary Sec.~\ref{sec:wrinkle}

We use the Adam optimizer to train all learnable parameters. When optimizing the Gaussian model, we inherit the learning rate adjustment from \cite{qian_gaussianavatars_2024} and apply the decaying strategy from the original 3DGS \cite{kerbl_3d_2023}. For efficiency, SDF, shape, appearance MLP, as well as feature grid are all implemented using adapted tinycudann \cite{muller_tiny-cuda-nn_2021} from $NeuS2$ \cite{wang_neus2_2023}, with a learning rate set at \(1\times {10}^{-4}\). We train the Gaussian model for 30,000 iterations, then we freeze the SDF branch. For the second training phase, the learning rates for the Gaussian attributes are kept unchanged. The entire process takes about 2 hours on an NVIDIA A100 GPU.

\subsection{Results and Comparisons}
\begin{figure}[t]
    \centering
    \captionsetup{type=figure}
    \includegraphics[width=\linewidth]{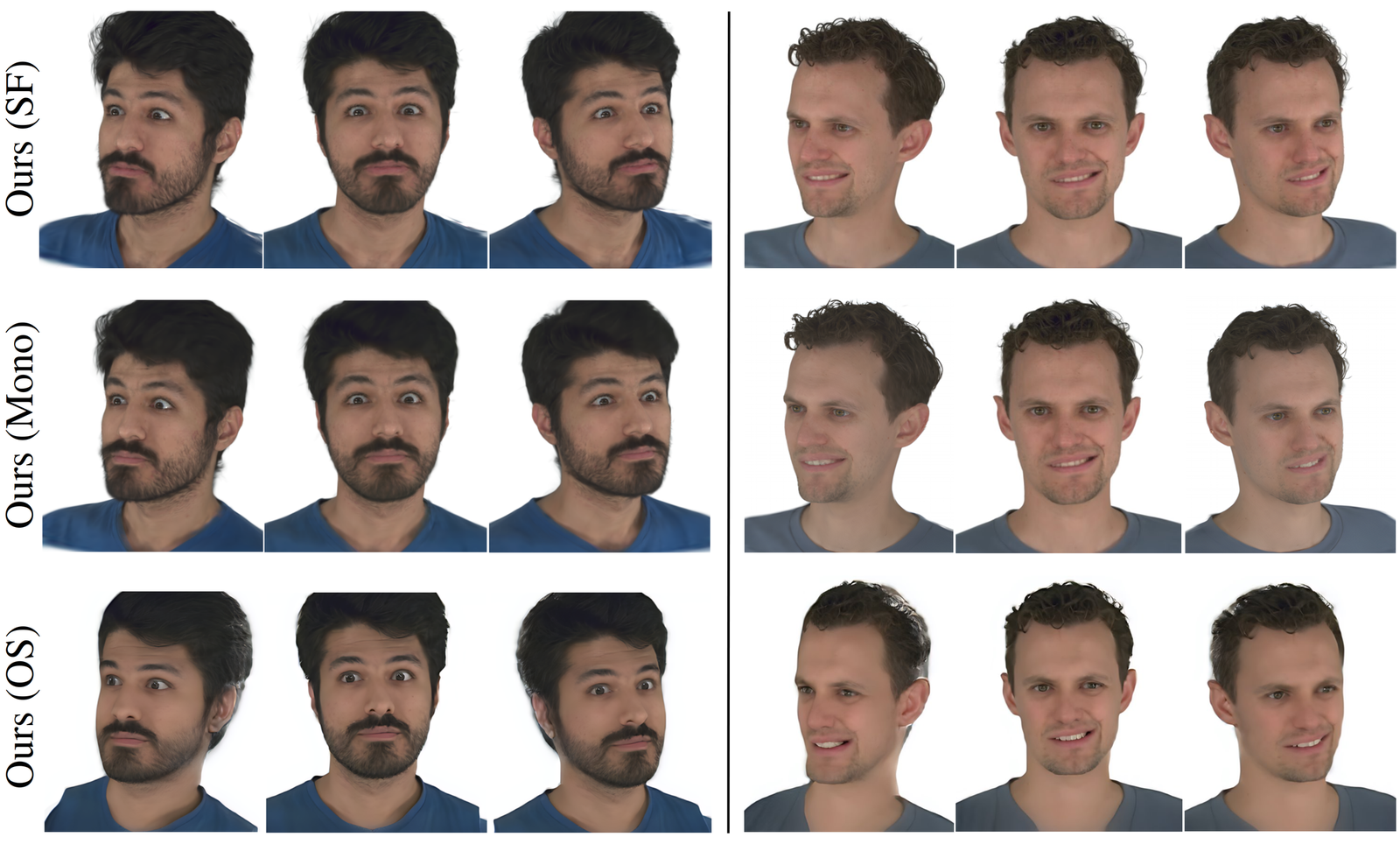}
    \caption{Qualitative evaluation of multi-view consistency of our pipeline in different reconstruction settings.}
    \label{fig:MV_new}
\end{figure}

The experiments are split into 5 groups based on input types. For the Multi-view reconstruction setting, we choose GaussianAvatars \cite{qian_gaussianavatars_2024} (GA) and Gaussian-Head-Avatar \cite{xu_gaussian_2024} (GHA). This group is further divided into 2 settings: reconstruction from the entire training sequence versus only taking 1 timestep. On full training data, GA takes about 5 hours and GHA takes 14 hours to train, significantly longer than our pipeline. In the one-shot setting, GPAvatar \cite{chu_gpavatar_2024} and LAM \cite{he_lam_2025} are used for comparison, taking only one frontal image as input. We also used MICA  tracking's FLAME expression parameter to drive animation for fair assessment (we kept the original translation and rotation parameters for pose accuracy). We use FlashAvatar \cite{xiang_flashavatar_2024} and RGBavatar \cite{li_rgbavatar_2025} as a comparison in the monocular setting. We also further divide this into training on the full training sequence versus taking only the head rotation sequence. The evaluation metrics include: Peak Signal-to-Noise Ratio (PSNR), Structure Similarity Index (SSIM), Learned Perceptual Image Patch Similarity (LPIPS) \cite{zhang_unreasonable_2018}. The results are reported in Tab. \ref{tab:main_tb} and Fig. \ref{fig:self-re}, a multi-view comparison between adaptations of our methods are shown in Fig. \ref{fig:MV_new}. Finally, we report the cross-reenactment quality in Fig. \ref{fig:four_by_six_new}. We report further qualitative results in supplementary material including geometrical quality assessment, see Sec.~\ref{sec:novel_view} and Sec.~\ref{sec:geo_results}

\begin{figure}[t]
    \centering
    \captionsetup{type=figure}
    \includegraphics[width=\linewidth]{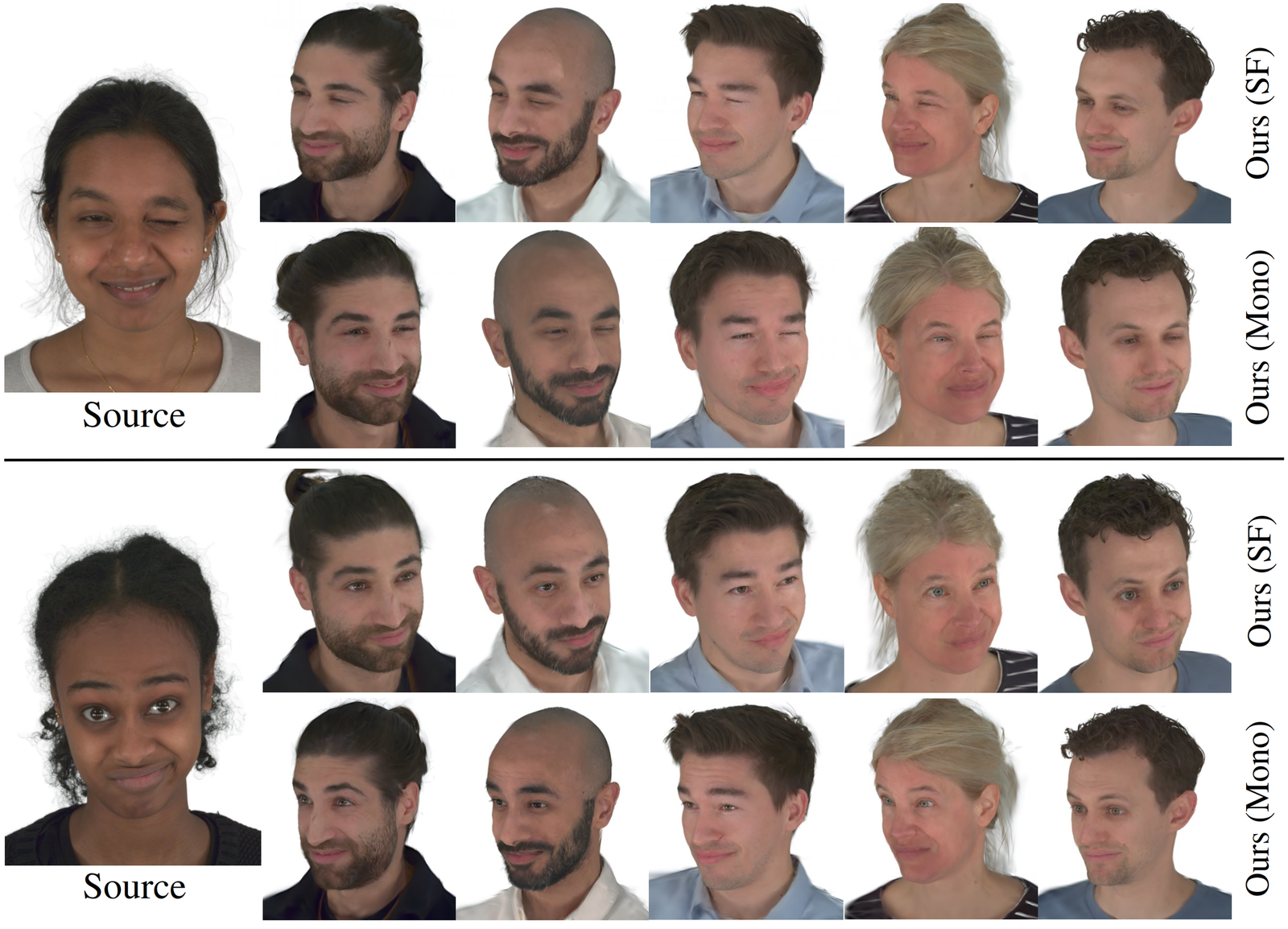}
    \caption{Qualitative comparisons for cross-reenactment. The images in the first column are the source avatars, and on the right are 2 adaptations of our method in 2 view angles.}
    \label{fig:four_by_six_new}
\end{figure}

\subsubsection{Self-Reenactment}
Self-reenactment serves as our primary evaluation task, as it directly reflects how well a reconstructed avatar can reproduce unseen expressions from the target identity. As summarized in Tab.~\ref{tab:main_tb}, our method achieves competitive or superior performance compared to state-of-the-art approaches across all reconstruction settings, despite not using explicit expression supervision. In multiview single-frame scenarios, baselines such as GA and GHA frequently exhibit texture distortions and geometry inconsistencies when reenacting novel expressions, whereas our approach produces more stable facial geometry and fewer artifacts.

Under the one-shot setting, GPAvatar and LAM struggle with geometric fidelity and exhibit noticeable degradation in facial details, often failing to maintain consistent appearance across view directions. In contrast, our one-shot variant preserves sharper identity features and achieves stronger multi-view consistency, leading to more coherent reenactments Fig.~\ref{fig:MV_new}. Similarly, in the monocular single-view setting, FlashAvatar and RGBavatar demonstrate siginificant drop in view consistency and failing to reconstruct geometry faithfully, resulting in muted expressions and view-dependent artifacts. Our pipeline, however, reconstructs avatars with more reliable geometry and better cross-view coherence, enabling accurate and stable expression transfer even from limited input observations. Additional qualitative and quantitative results, including further comparisons across view directions, are provided in the supplementary material Sec.~\ref{sec:novel_view}.

We further analyze GA and GHA's robustness to limited training data, the quality of these baseline reconstructions improves steeply as additional frames are introduced, revealing a strong dependence on dense expression supervision, please refer to supplementary Sec.~\ref{sec:dependency} and Fig.~\ref{fig:short}.

\subsection{Ablation Study}

\begin{table}
    \centering
    \small  
    \setlength{\tabcolsep}{2pt} 
    \renewcommand{\arraystretch}{1.1} 

    \scalebox{0.86}{
    \begin{tabular}{l|ccc|ccc}
        \toprule
        &  \multicolumn{3}{c|}{\textbf{Novel-View}} & \multicolumn{3}{c}{\textbf{Self-Reenactment}} \\
        & PSNR↑ & SSIM↑ & LPIPS↓ & PSNR↑ & SSIM↑ & LPIPS↓ \\
        \midrule
        SAGE &33.79 & 0.963 & 0.071  & 26.64 & 0.907 & 0.093 \\
        w/o \(MLPs\)& 33.95 &0.961 & 0.071 & 24.72 & 0.857 &  0.142\\
        w/o $2^{nd}$ phase& 34.14 & 0.962 & 0.067 &  22.18 & 0.800 & 0.132 \\
        w/o Correction& \multicolumn{3}{c|}{\textbf{Same as full}} & 25.42 & 0.905 & 0.101 \\
        \bottomrule
    \end{tabular}
    }
    \caption{Ablation study conducted on Novel-View and Self-Reenactment qualities, ablating Gaussian attributes Parametrization (w/o MLPs); self-supervision phase; and Gaussian Surfel Correction. This is a harder target for accurate geometric reconstruction due to rich facial hair.}
    \label{tab:performance}
\end{table}

\begin{figure}
    \centering
    \captionsetup{type=figure}
    \includegraphics[width=\linewidth]{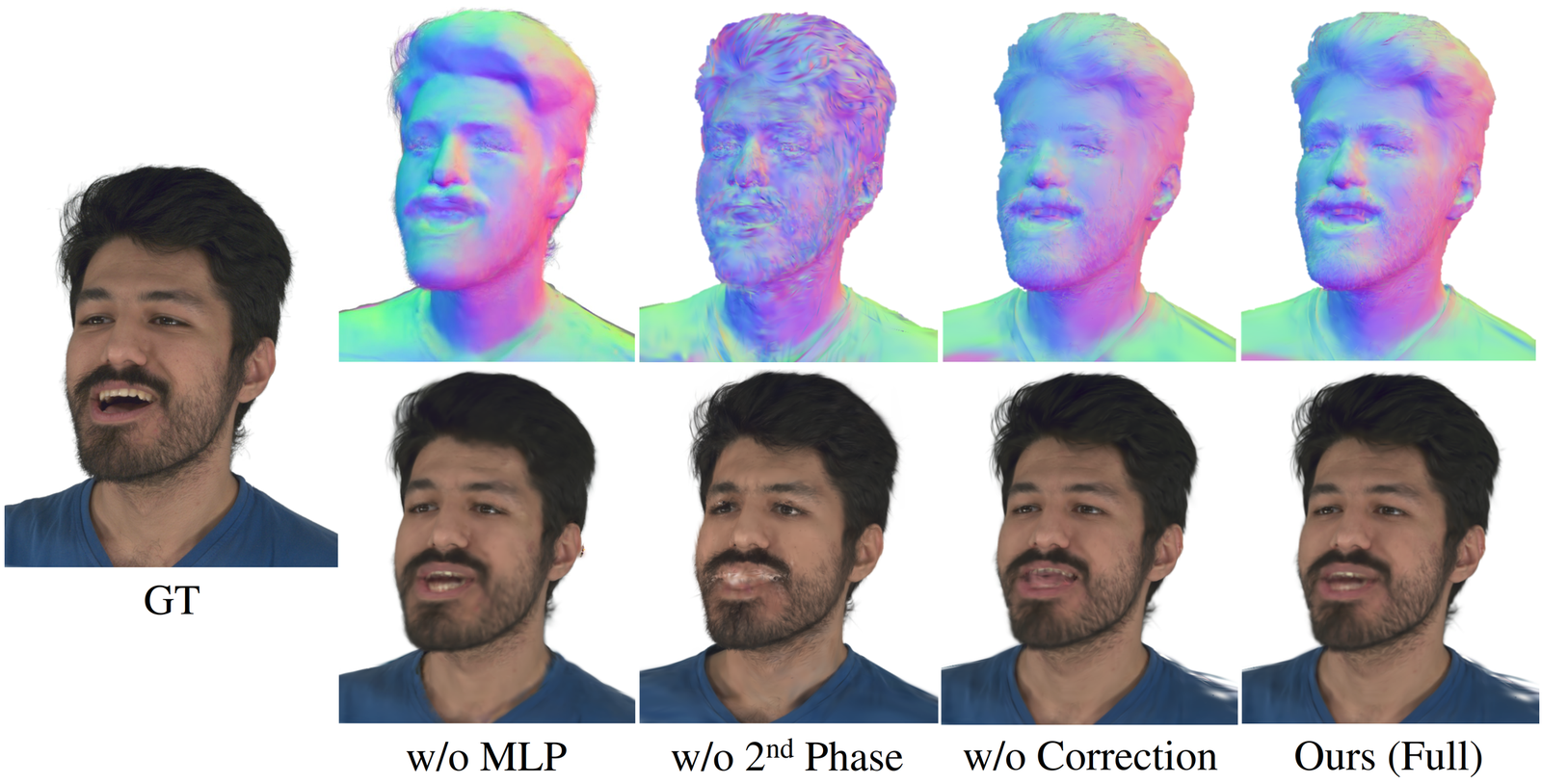}
    \caption{Qualitative comparison of geometry and rendered appearance ablated settings.}
    \label{fig:ablation_normal}
\end{figure}

\textbf{SDF.} The SDF serves as a core building block within our pipeline, providing geometric feature vectors to the appearance network as well as providing surface properties during Gaussian shape adjustment. 2DGS's constraints alone cannot effectively constrain surfel behaviour in a dynamic scene. Since the rest of our novel features heavily depends on the SDF, we do not ablate the joint optimization feature.

\noindent \textbf{Parametrizing Gaussian attributes.}
The removal of the MLPs negatively effect the fine-grained details of the rendered avatar, since low level geometric clue from SDF cannot be passed to individual Gaussians. In terms of geometrical quality omparing to our full model, the ablated version show less detailed geometry in facial hair regions and eye regions, and is closer to the coarse FLAME fitting. In terms of visual quality, Gaussian colors become more randomized, and without a canonical representation, we cannot impose appearance supervision in the second training phase, leading to noisier and potentially blurry reenactment results. 

\noindent \textbf{Self-Supervision Phase.}
Our pipeline heavily depends on the second training phase to enforce geometrical consistency during reenactment, a large performance drop can be observed from Tab. \ref{tab:performance} and noisy geometries and significant artifact is shown in Fig. \ref{fig:ablation_normal}. Interestingly, this training phase has a negative impact on NVS quality, but an increase in the reenactment task, indicating an obvious trade-off between generalizability to deformation or reconstruction quality in static scenes.

\noindent \textbf{Shape correction.}
Disabling shape correction yield a slight degradation in both visual and geometry quality. In Fig. \ref{fig:ablation_normal} we see a less defined brow bone and some artifact can be spotted in mouth region. Meanwhile, without the shape correction feature, the facial geometry could be over-smoothed when enforcing distortion and normal consistency loss during self-supervision phase. Further detail can be found in supplementary Sec.~\ref{sec:novel_view}

\subsection{Limitations and Future Work} Our method's accuracy depends on the expressiveness and correctness of the FLAME parameters, as expressions that cannot be modeled by FLAME tracking produce inaccuracies induced into our rendered results, causing deviations from the ground truth. Additionally, unobserved regions relies on pre-fitting such as teeth and tongue.

However, our pipeline design is very flexible and can easily be incorporated into any other Gaussian based non-rigid object animations such as human body, as long as a per Gaussian transformation can be inferred.  The adaptations we demonstrated are not optimal; for example, we could instead use a normal map conditioned diffusion \cite{tang_gaf_2025, kirschstein_diffusionavatars_2024} to generate pseudo multiview with higher accuracy. As a future direction, we could substitute FLAME parametrized 3DMM for NPHM, which is more expressive. Furthermore, a synthetic prior with identity encoding could be introduced for even better few-shot reconstruction performance.
\vspace{-0.2cm}
\section{Discussion and Conclusion}
\textbf{Ethical Considerations.} The creation of animatable photorealistic avatars raises concerns about privacy, consent, and misuse. Ensuring explicit user consent, securing biometric data, and preventing unauthorized use are essential. Risks like identity theft, deepfakes, and bias must be mitigated through security measures, watermarking, and diverse datasets. Ethical AI practices and transparency should guide responsible deployment.

\noindent \textbf{Conclusion.}We present SAGE, a pipeline for reconstructing animatable Gaussian avatars from minimal expression data by enforcing geometric constraints for transformation consistency. To address Gaussian distribution, we optimize SDF for surface supervision, introduce shape corrections, and parametrize Gaussian attributes. A self-supervision strategy is proposed to replace long training sequences, drastically reducing data requirements. SAGE achieves SOTA performance with minimal data, which could be applied to real-life applications more efficiently.

{\small
\bibliographystyle{ieee}
\bibliography{egbib}
}

\clearpage
\setcounter{page}{1}

\maketitlesupplementary
\section{Shape Network}\label{sec:shape_net}
We design the shape network to predict the lower-triangular part of a matrix and a scaler value, the lower triangle is then assembled into a symmetric Hessian. Eigen-decomposition of this matrix yields three eigenvalue–eigenvector pairs. The eigenvector corresponding to the smallest eigenvalue is interpreted as the surfel normal; we explicitly enforce this eigenvalue to be close to zero and use the associated vector in Equation~\ref{eq:7}. The remaining two eigenvectors span the tangent plane and serve as the basis for the rotational components. Intuitively, the network is learning to predict the local tangent plane of the SDF at a query point. We rescale the two non-zero eigenvalues by dividing them by the larger of the pair, and then scale them using an additional scalar predicted by the network. The resulting two values are swapped then used as the surfel scales in global scale. This formulation naturally aligns with Section~\ref{sec:correction}, where stretching and curvature changes are computed within the tangent space.

\begin{figure}[!t]
    \centering
    \includegraphics[width=0.9\linewidth]{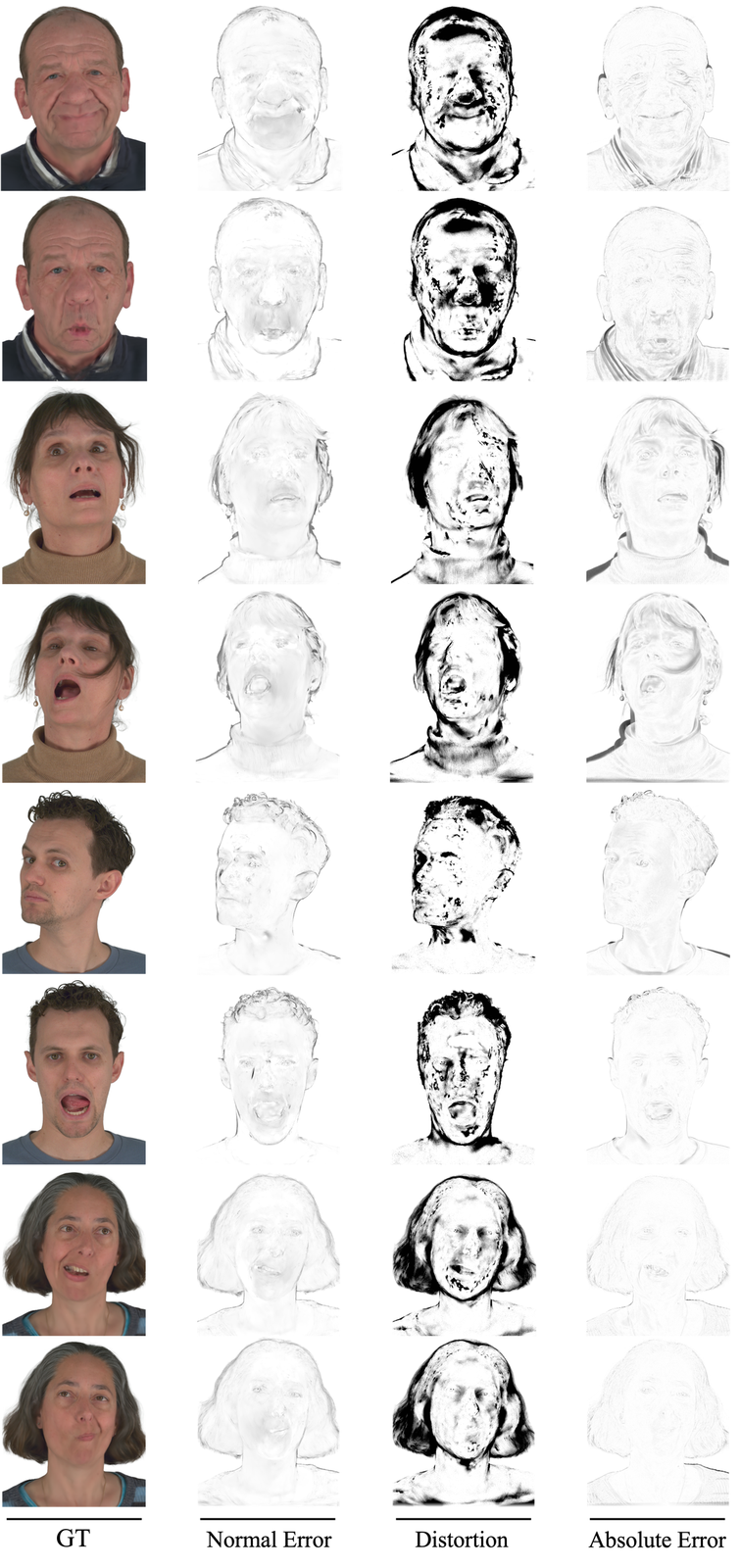}
    \caption{We visualize three pixel-level error maps used in our 2DGS optimization pipeline—normal error, distortion error, and the absolute difference between our reconstruction and ground truth. The results reveal a strong correlation between geometric inaccuracies, regions of higher reconstruction difficulty, and the presence of expression-dependent facial creases.}
    \label{fig:wrinkle}
\end{figure}
\section{Dynamic Wrinkles}\label{sec:wrinkle}
We observed a strong correlation between normal-estimation error and  the appearance of wrinkles, as well as between distortion errors and local surface-orientation changes. As shown in the reference image Fig.~\ref{fig:wrinkle}, regions showing large normal variation tend to coincide with fine-scale facial creases and expression-dependent wrinkles. Similarly, regions showing high distortion also feature dynamic changes in surface orientation, which can naturally lead to subtle view-dependent shading. Additionally, reconstruction loss are higher in those regions. Building off of these observations, we darken Gaussians with high normal error, which increases wrinkle visibility and realism, while dimming them slightly in high-distortion areas to obtain a representation of dynamic shadowing and orientation-based shading. Furthermore, during the second training phase, we focus sampling in these regions to compare with canonical reconstruction. This strategy allows for fine-grained facial effects to be created within the reconstructed avatars in an efficient, lightweight environment.

\section{Monocular Reconstruction}\label{sec:mono_recon}
\begin{figure}[h]
    \centering
    \includegraphics[width=\linewidth]{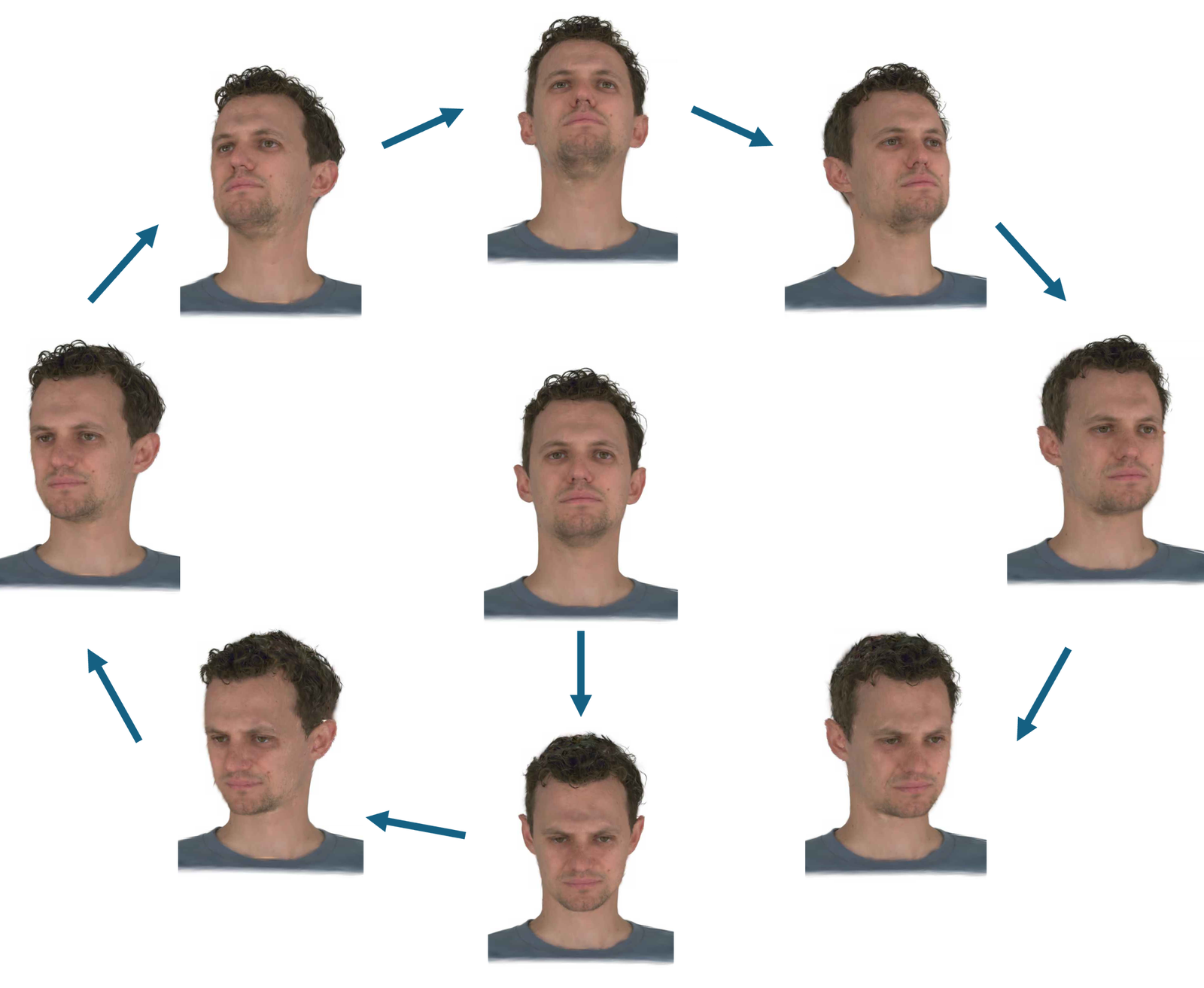}
    \caption{Visualization of the curated monocular dataset. We generate synthetic head rotation sequences by adjusting only rotation parameter. Shown are representative frames illustrating continuous head rotations.}
    \label{fig:monodata}
\end{figure}

Since no existing human head dataset provides a clean separation between head rotation and expression sequences, we curate a synthetic sequence of FLAME parameters where only head rotations vary while all expression parameters remain fixed. This sequence is reenacted with a pre-trained GaussianAvatar~\cite{yu_one2avatar_2024} model to generate synthetic head rotation data. In total, we synthesize approximately 40 frames of head rotations and complement them with around 20 real head rotation frames from the original dataset, forming the SV-HR subset used in Tab.~\ref{tab:main_tb}. We demonstrate this dataset using a selection of frames in Fig.~\ref{fig:monodata}.

We then employ MonoNPHM’s tracker to process this data, leveraging its deformation module to accurately warp rays from dynamic scenes to the canonical space during the first training phase. Specifically, we adopt the globalMLP variant of their pre-trained models, since we do not utilize their local SDF ensemble to represent head geometry. For simplicity, we also omit their hyper-dimensional extensions.

\section{Novel View Synthesis Results}\label{sec:novel_view}
We do not report numerical NVS metrics because our experiments are organized into five input-type groups, each containing distinct sets of novel expressions and viewpoints, making direct quantitative comparison infeasible. Instead, we present a qualitative assessment of multi-view consistency under sparse-view conditions, comparing our method with RGBAvatar and LAM. As shown in Figure X, our pipeline delivers noticeably stronger cross-view consistency under identical reconstruction settings, producing coherent appearance and geometry across viewpoints where baseline methods exhibit view-dependent artifacts.

\section{Geometric Reconstruction Results}\label{sec:geo_results}
\begin{figure*}[t]
    \centering
    \includegraphics[width=0.8\linewidth]{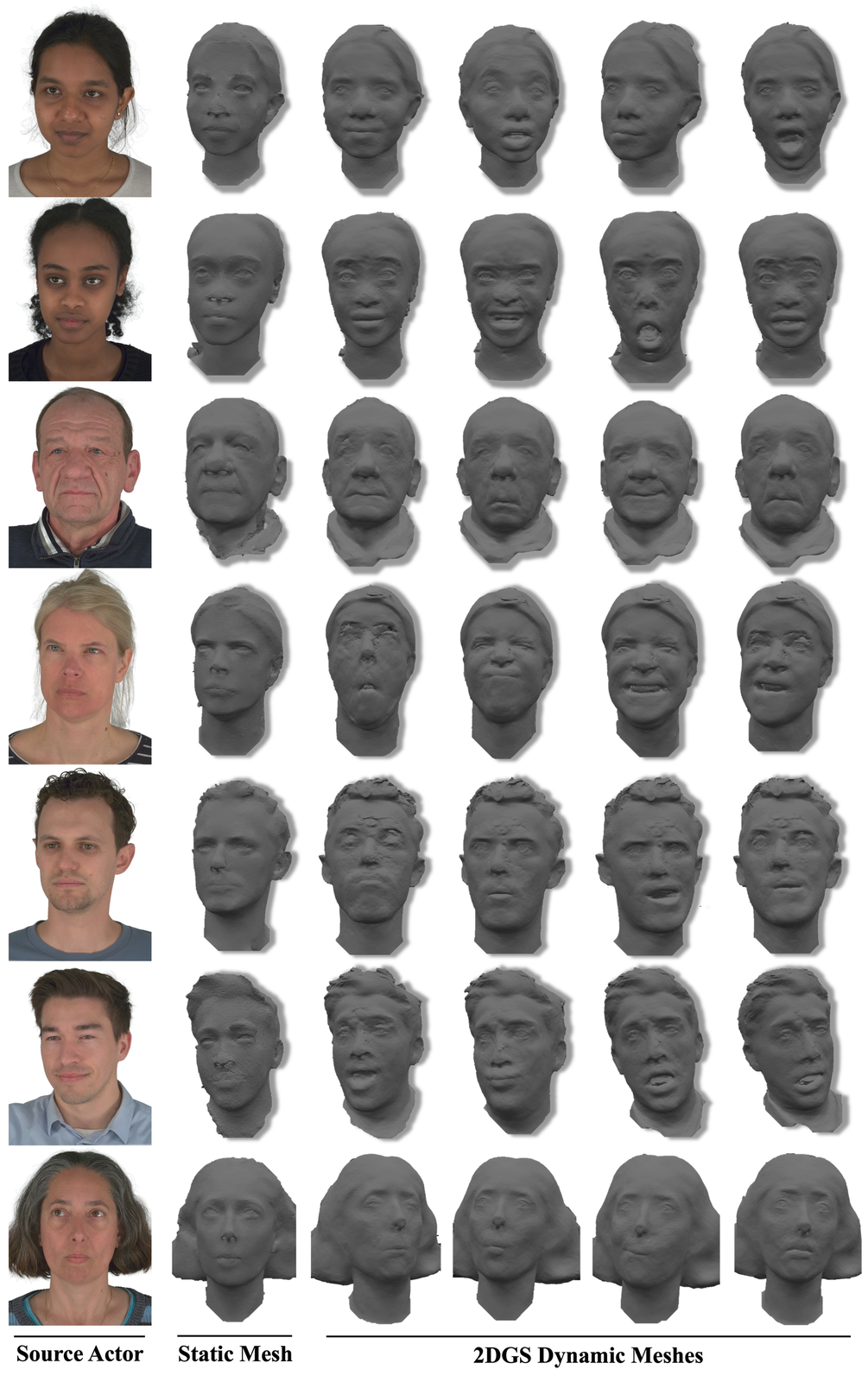}
    \caption{Static canonical SDF meshes and reconstructed dynamic meshes across identities and expressions. The canonical SDF captures clean, identity-preserving geometry, while the dynamic reconstructions accurately model a wide range of expression-dependent deformations.}
    \label{fig:mesh_grid}
\end{figure*}

We demonstrate our first stage of joint optimization reconstruct accurate geometry as shown in Fig.~\ref{fig:mesh_grid}, capturing identity-specific facial structure with clean surface topology and stable normals. When generalized to dynamic sequences, canonical SDF’s well-defined geometry acts as a strong prior that constrains and stabilizes dynamic surface deformation, preventing drift and artifacts. We also present the retrieved dynamic meshes during self-reenactment, the geometric accuracy can be faithfully transferred to the dynamic settings across subjects and expressions, proving the effectiveness of our pipeline.

\section{Gaussian Correction}\label{sec:gaus_correction}
\begin{figure}[h]
  \centering
  \includegraphics[width=0.8\linewidth]{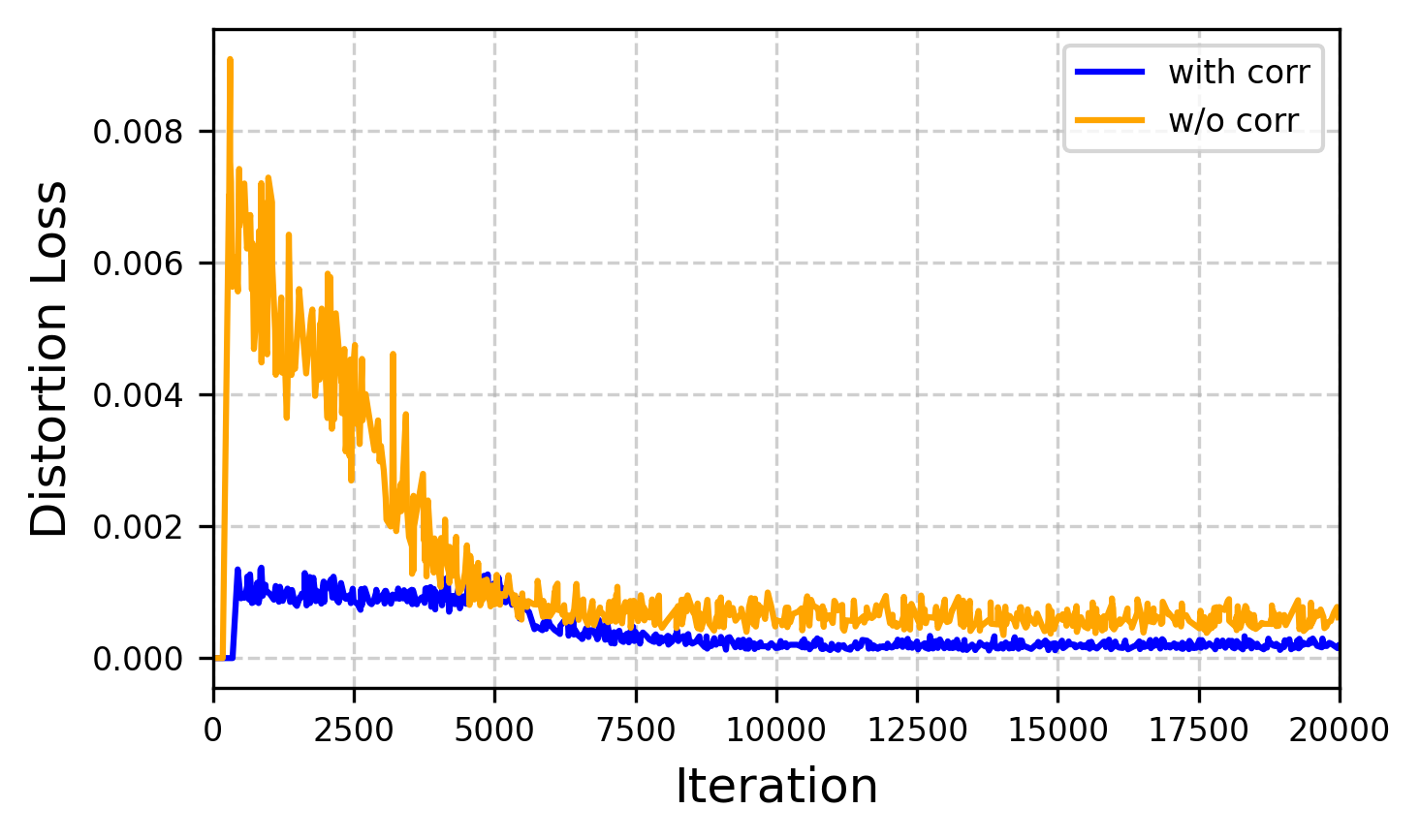}
   \caption{Training log of distortion loss on the second training phase. It is clear that with correction, distortion starts from a much lower value and less volatile initially.}
   \label{fig:dist}
\end{figure}

We demonstrate the effectiveness of our Gaussian correction strategy in Fig.~\ref{fig:dist}, which illustrates the optimization of the distortion loss during the self-supervised training phase. With correction enabled, the distortion loss starts at a lower value and converges to a lower minimum, indicating improved stability. This shows that the corrected Gaussians align more closely with the deformed geometry, while also reducing overlap and collisions between individual Gaussians under deformation.\\

\begin{figure*}[t]
    \centering
    \includegraphics[width=0.8\linewidth]{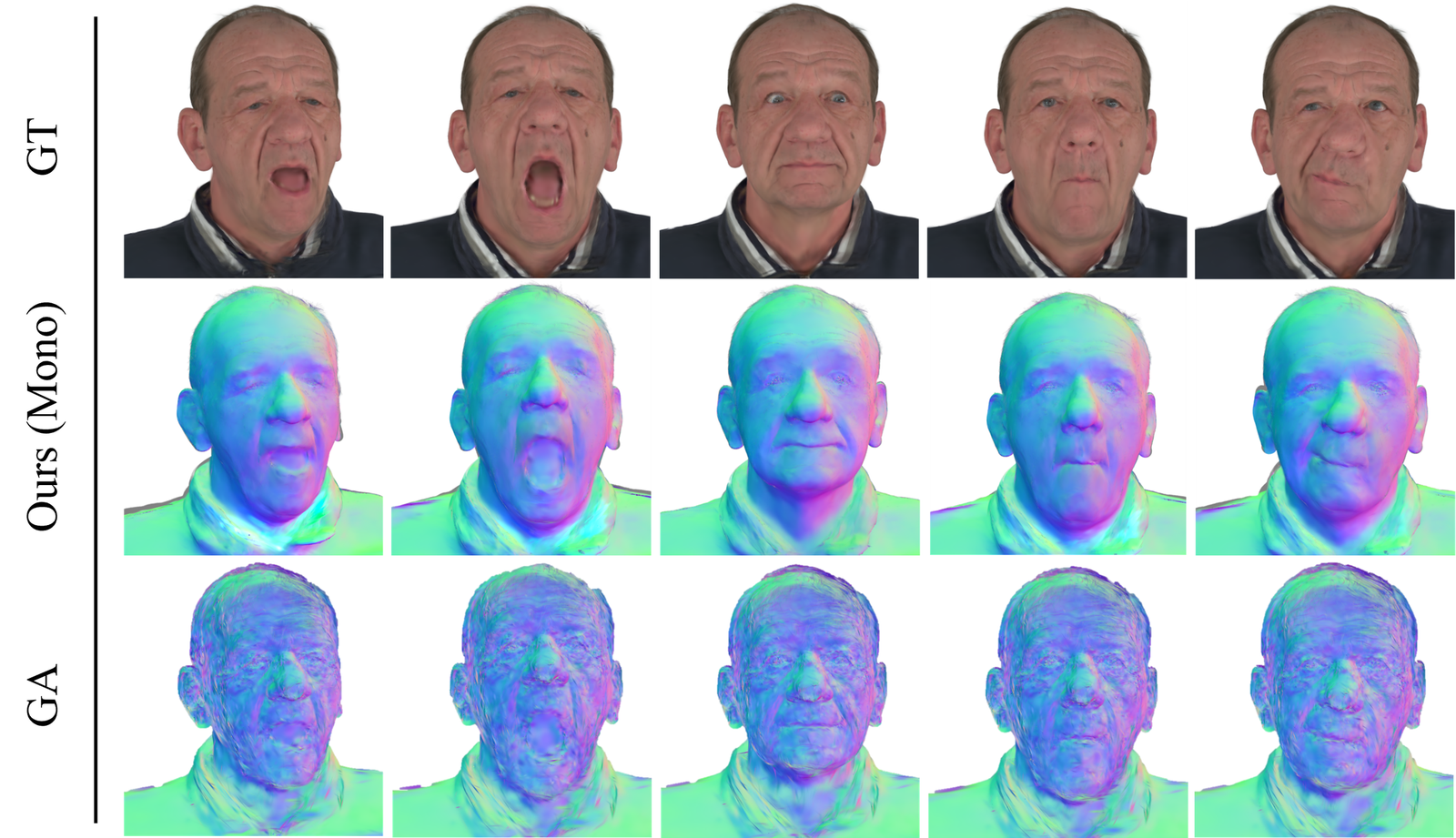}
    \caption{Comparison of reconstructed normals across a range of facial expressions. Middle row is the monocular adaptation of our pipeline, the bottom is GA but adapted with 2DGS}
    \label{fig:210_grid}
\end{figure*}

\begin{figure*}[t]
    \centering
    \includegraphics[width=0.8\linewidth]{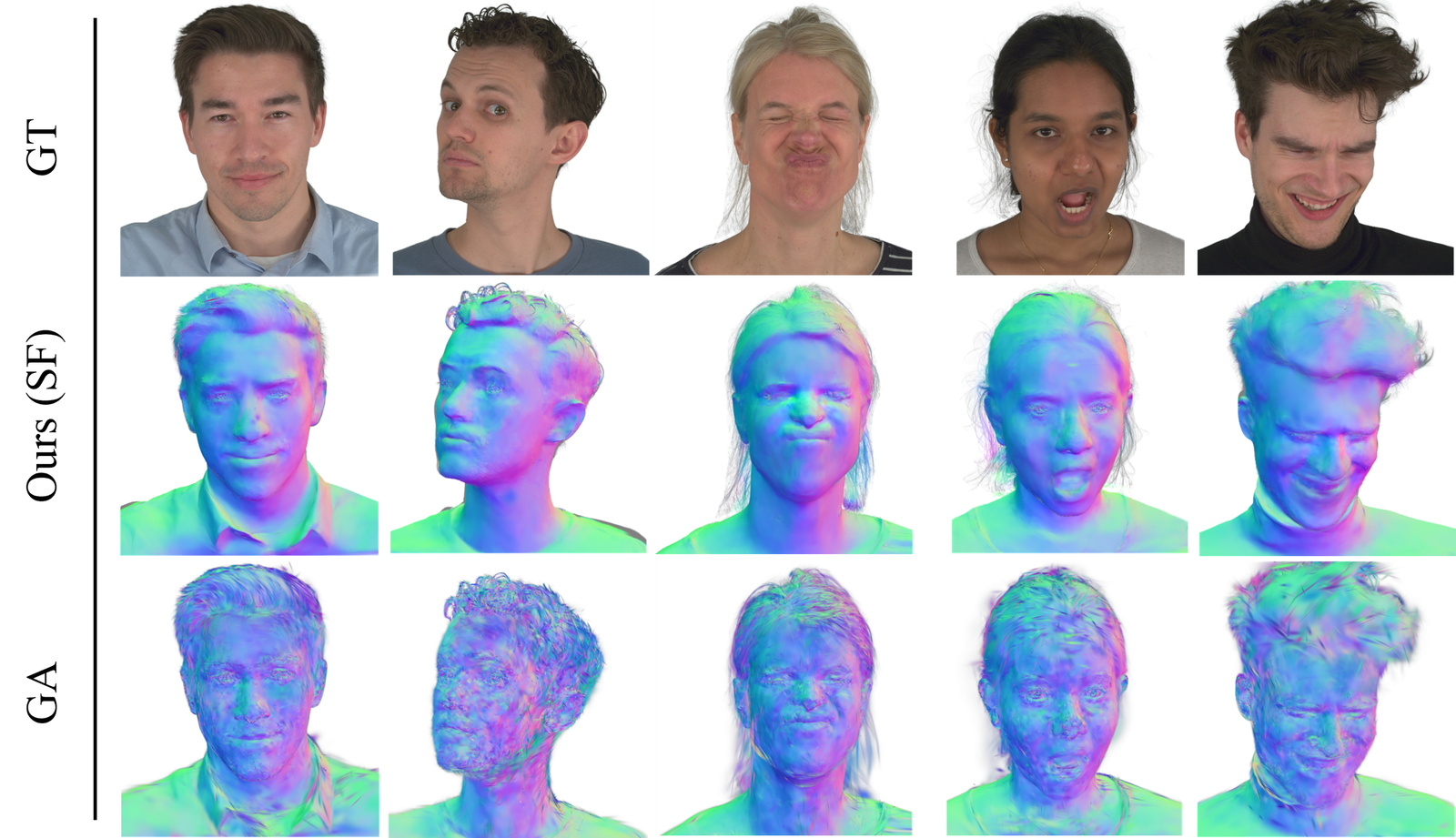}
    \caption{Comparison of reconstructed normals across a range of identities. Middle row is the multiview adaptation of our pipeline, the bottom is GA but adapted with 2DGS.}
    \label{fig:multi_grid}
\end{figure*}

\section{Expression Data Dependency}\label{sec:dependency}
In this section, we evaluate the robustness of our pipeline from two perspectives: 
(1) increasing the number of training expressions to assess the method's dependency on expression supervision, and 
(2) varying the expression chosen for canonical reconstruction in the first stage to examine its impact on expression generalization.
\begin{figure*}[t]
    \centering
  \begin{subfigure}[b]{0.3\linewidth}
    \includegraphics[width=\linewidth]{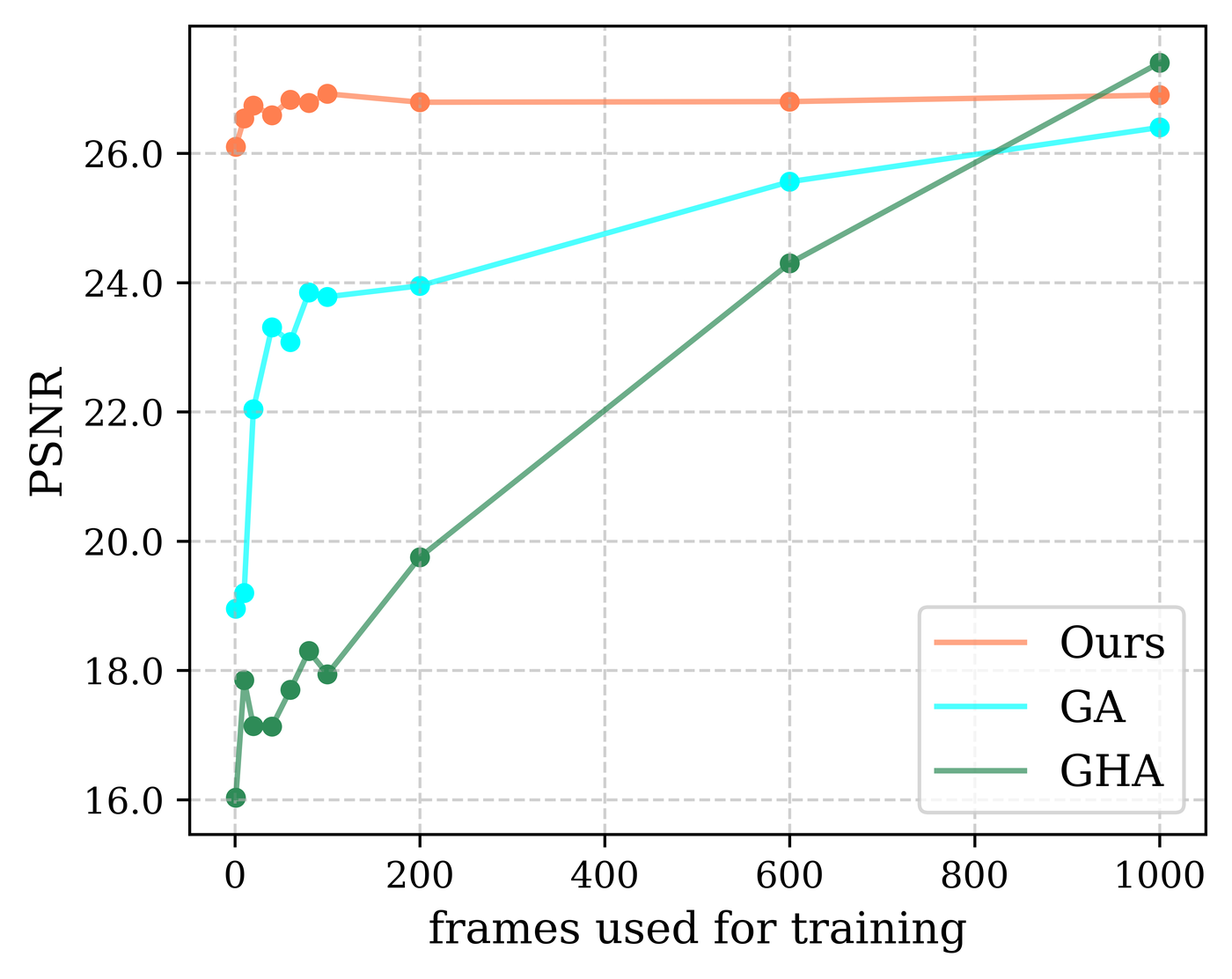}
  \end{subfigure}
  \begin{subfigure}[b]{0.3\linewidth}
    \includegraphics[width=\linewidth]{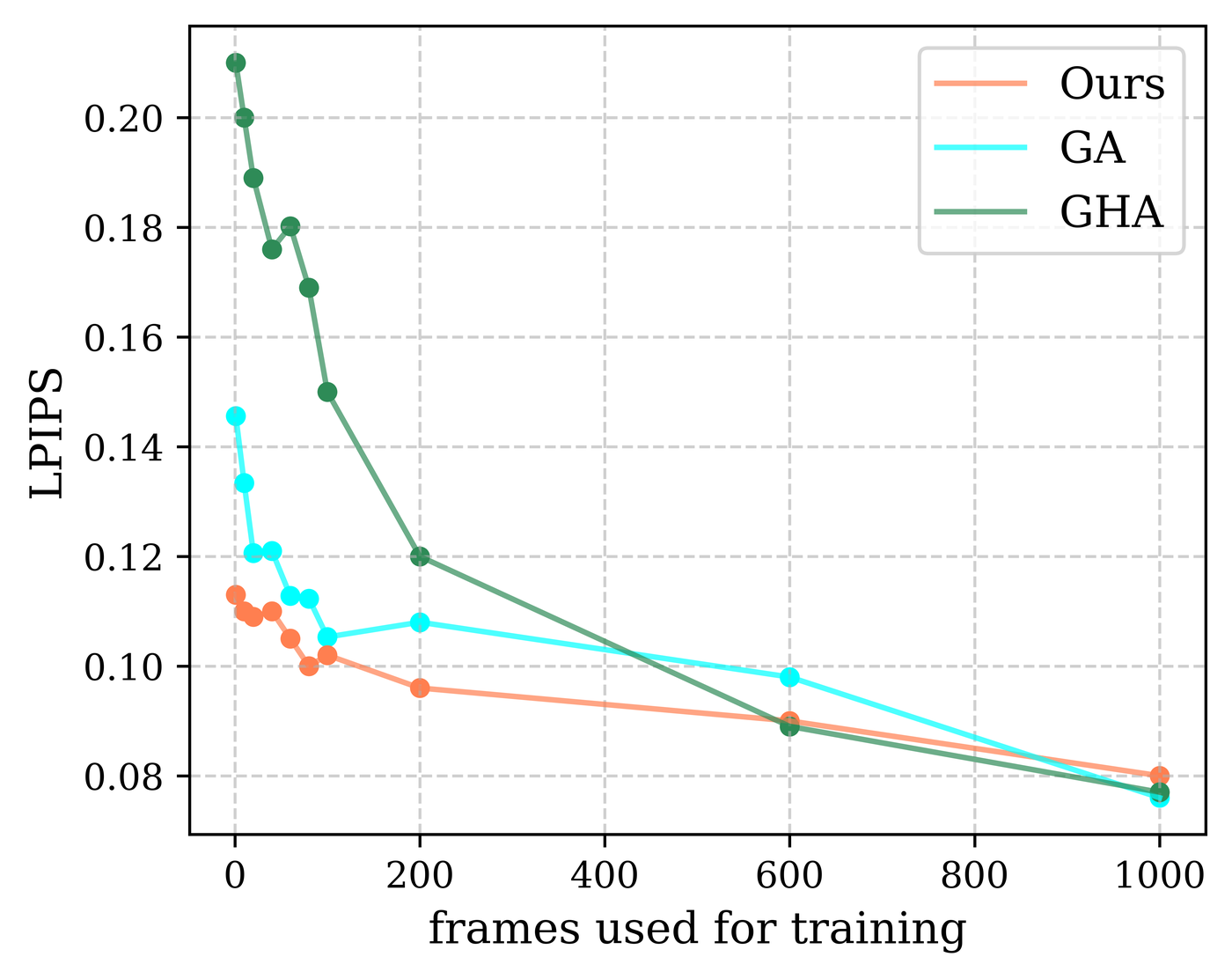}
  \end{subfigure}
  \begin{subfigure}[b]{0.3\linewidth}
    \includegraphics[width=\linewidth]{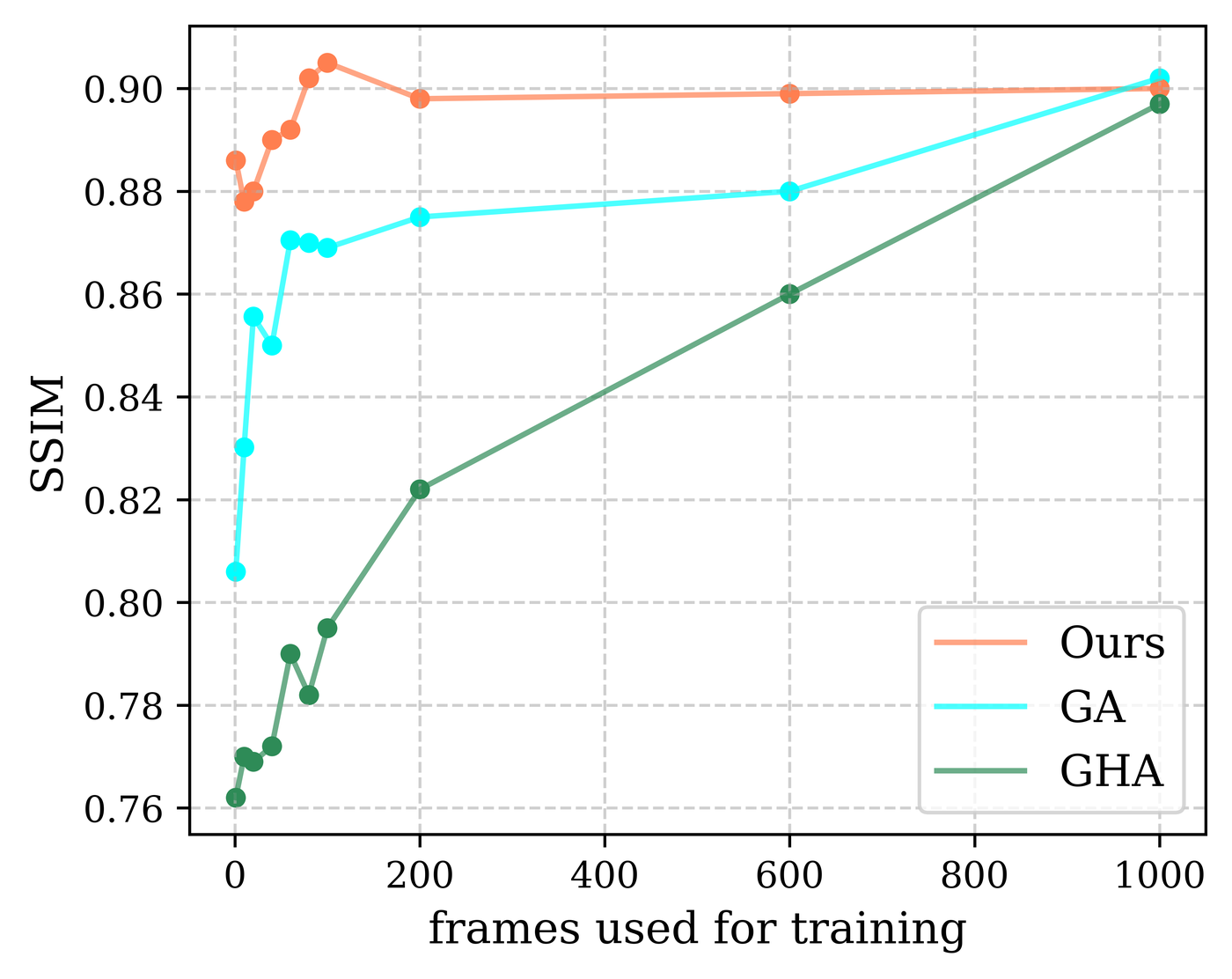}
  \end{subfigure}
  \caption{Experiment on baseline models' self-reenactment quality as the number of frames used for training increases.}
  \label{fig:short}
\end{figure*}

We evaluate the first aspect under multiview setting reconstructing participant 074. Our pipeline demonstrates strong robustness across varying amounts of training data, as shown in Fig.~\ref{fig:short}. Under expression sparse scenario, our pipeline achieves consistently high reconstruction quality, outperforming baseline methods in PSNR, LPIPS, and SSIM. As the number of training frames increases, our method maintains stable performance with only marginal fluctuations, indicating that it does not rely heavily on dense temporal supervision to achieve high fidelity.

\begin{figure*}
    \centering
    \includegraphics[width=0.9\linewidth]{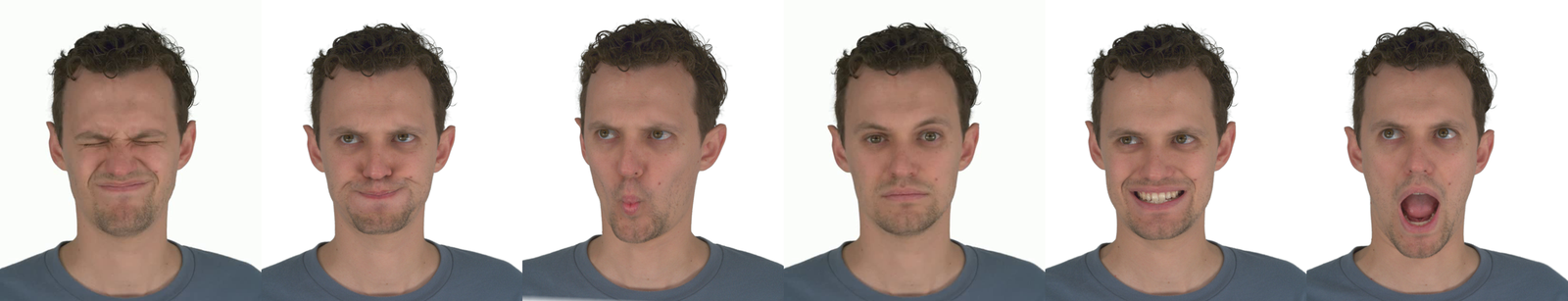}
    \caption{Visualization of the six expressions used in the expression-dependency study. From left to right, the selected expressions range from eyes closed (exp-1) to mouth open (exp-6), covering a diverse set of facial configurations.}
    \label{fig:306}
\end{figure*}

For the second perspective, we reconstruct the canonical SDF for participant 306 using six distinct expressions (Fig.~\ref{fig:306}), ranging from eyes-closed (exp-1) to mouth-open (exp-6), while keeping the second training phase fixed. We perform this evaluation in both monocular and multiview settings and report the averaged self-reenactment metrics in Tab.~\ref{tab:ED}. The results vary only slightly across canonical choices, indicating that our pipeline is largely insensitive to which expression is selected as the canonical reference. Although a neutral expression with opened mouth tend to produce marginally higher fidelity, the overall differences remain small. A noticeable performance drop is observed for more extreme expressions such as exp-1, where the lack of visible eyeballs and the concentration of fine facial creases introduce additional reconstruction challenges, resulting in reduced generalization quality.

\begin{table}[h!]
\centering
\begin{tabular}{c|c|c|c}
\hline 
Expression & PSNR ↑ & SSIM ↑ & LPIPS↓ \\
\hline
1 & 25.62 & 0.873 & 0.118 \\
2 & 26.94 & 0.885 & 0.109 \\
3 & 26.53 & 0.868 & 0.123 \\
4 & 27.31 & 0.892 & 0.102 \\
5 & 27.12 & 0.906 & 0.110 \\
6 & 27.32 & 0.890 & 0.099 \\
\hline
\end{tabular}
\caption{Quantitative evaluation of expression generalization when varying the canonical expression used for SDF reconstruction.}
\label{tab:ED}
\end{table}

\begin{figure*}[t]
    \centering
  \begin{subfigure}[b]{0.85\linewidth}
    \includegraphics[width=\linewidth]{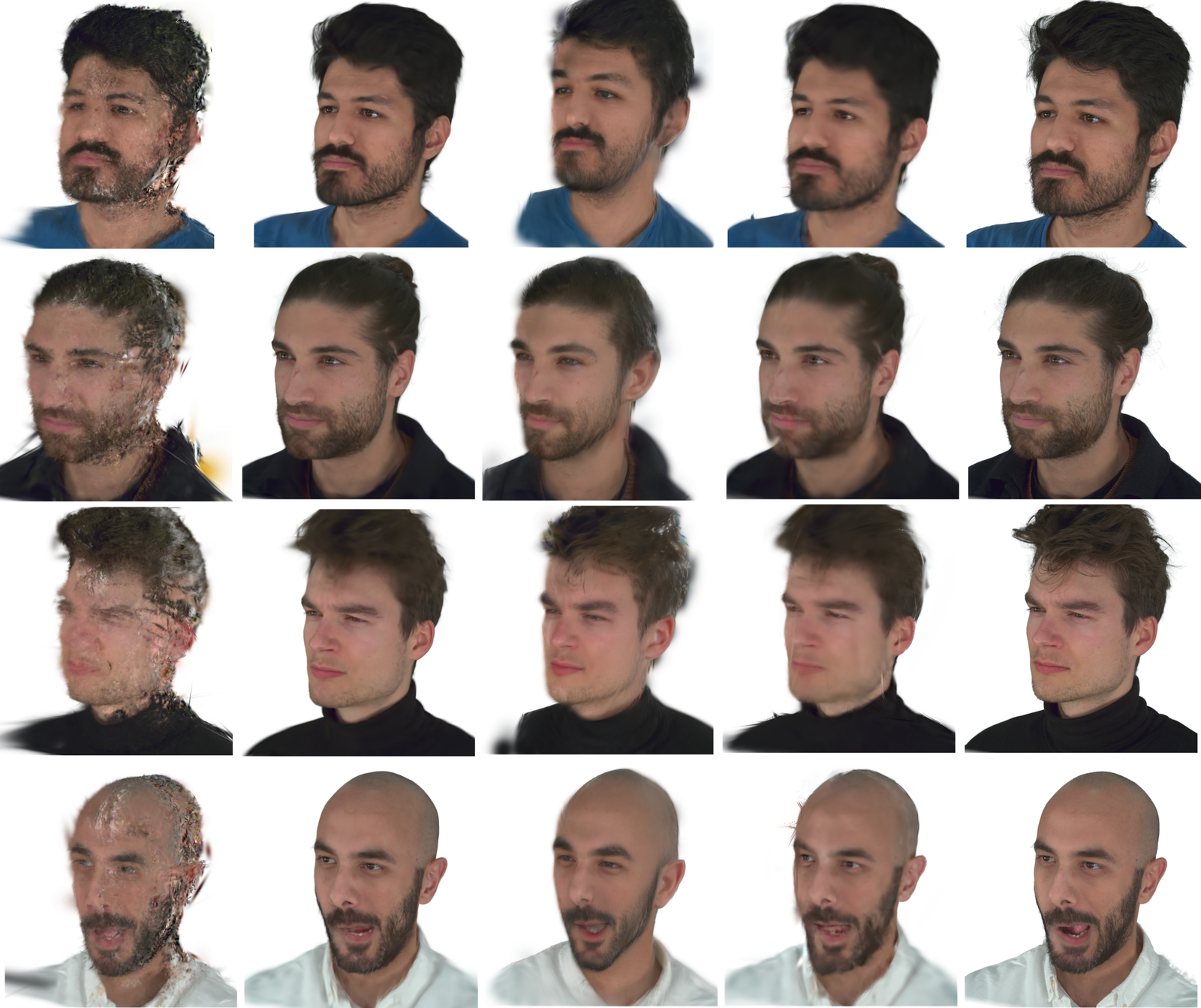}
  \end{subfigure}
  \begin{subfigure}[b]{0.85\linewidth}
    \includegraphics[width=\linewidth]{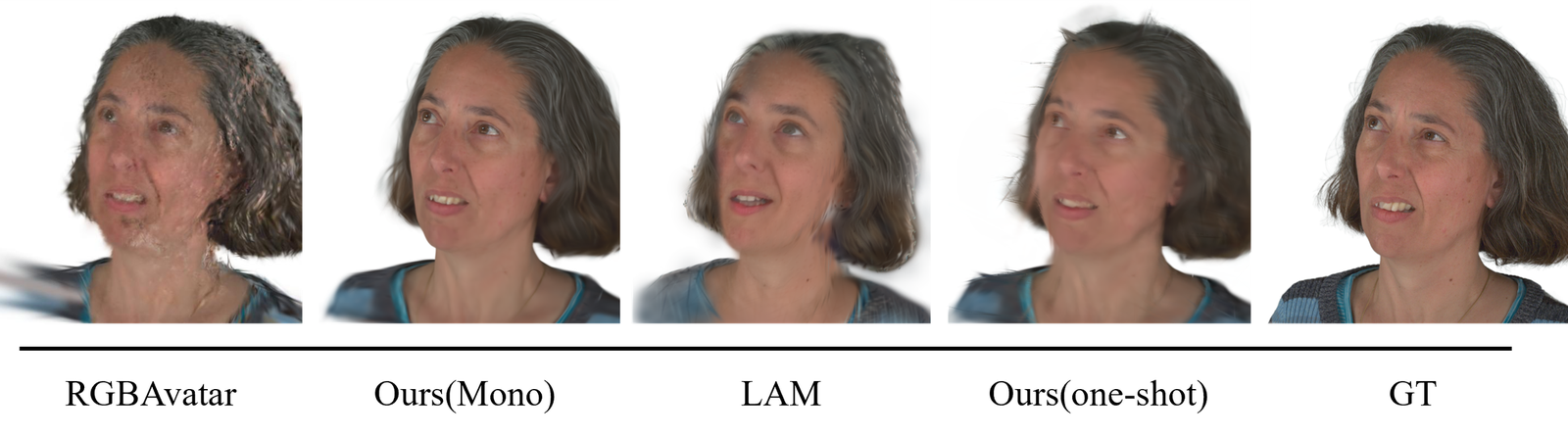}
  \end{subfigure}
  \caption{Qualitative comparison of novel-view synthesis under sparse-view input conditions. We compare RGBAvatar, LAM, our monocular setting (Ours(Mono)), and our one-shot setting (Ours(one-shot)) against ground truth. Baseline methods often exhibit view-dependent artifacts, blurring, or geometry–appearance inconsistencies,}
  \label{fig:multiview_grid}
\end{figure*}

\end{document}